\documentclass[10pt,twocolumn,letterpaper]{article}

\usepackage{cvpr}
\usepackage{times}
\usepackage{epsfig}
\usepackage{graphicx}
\usepackage{amsmath}
\usepackage{amssymb}
\usepackage{float}
\usepackage{subcaption} 

\usepackage{draftwatermark}
\SetWatermarkText{*** Accepted at CVPR 2020 ***}
\SetWatermarkScale{0.15}
\SetWatermarkAngle{0}
\SetWatermarkVerCenter{30px}

\usepackage[breaklinks=true,bookmarks=false]{hyperref}

\cvprfinalcopy 

\def\cvprPaperID{5427} 

\ifcvprfinal\fi

\begin{document}

\title{Watch your Up-Convolution: CNN Based Generative Deep Neural Networks are Failing to Reproduce Spectral Distributions}

\author{Ricard Durall$^{1,3}$, Margret Keuper$^{2}$, Janis Keuper$^{1,4}$\\[5mm]
$^1$Competence Center High Performance Computing, Fraunhofer ITWM, Kaiserslautern, Germany\\
$^2$Data- and Webscience Group, University Mannheim, Germany\\
$^3$IWR, University of Heidelberg, Germany\\
$^4$Institute for Machine Learning and Analytics, Offenburg University, Germany
}

\maketitle

\begin{abstract}
   Generative convolutional deep neural networks, \eg popular GAN architectures, are relying on convolution based up-sampling methods to produce non-scalar outputs like images or video sequences. In this paper, we show that common up-sampling methods, \ie known as \textit{up-convolution} or \textit{transposed convolution}, are causing the inability of such models to reproduce spectral distributions of natural training data correctly. This effect is independent of the underlying architecture and we show that it can be used to easily detect generated data like \textit{deepfakes} with up to 100\% accuracy on public benchmarks.
   To overcome this drawback of current generative models, we propose to add a novel spectral regularization term to the training optimization objective. We show that this approach not only allows to train spectral consistent GANs that are avoiding high frequency errors. Also, we show that a correct approximation of the frequency spectrum has positive effects on the training stability and output quality of generative networks. 
\end{abstract}

\section{Introduction}

\begin{figure}[t]
\centering
   \includegraphics[width=0.9\linewidth]{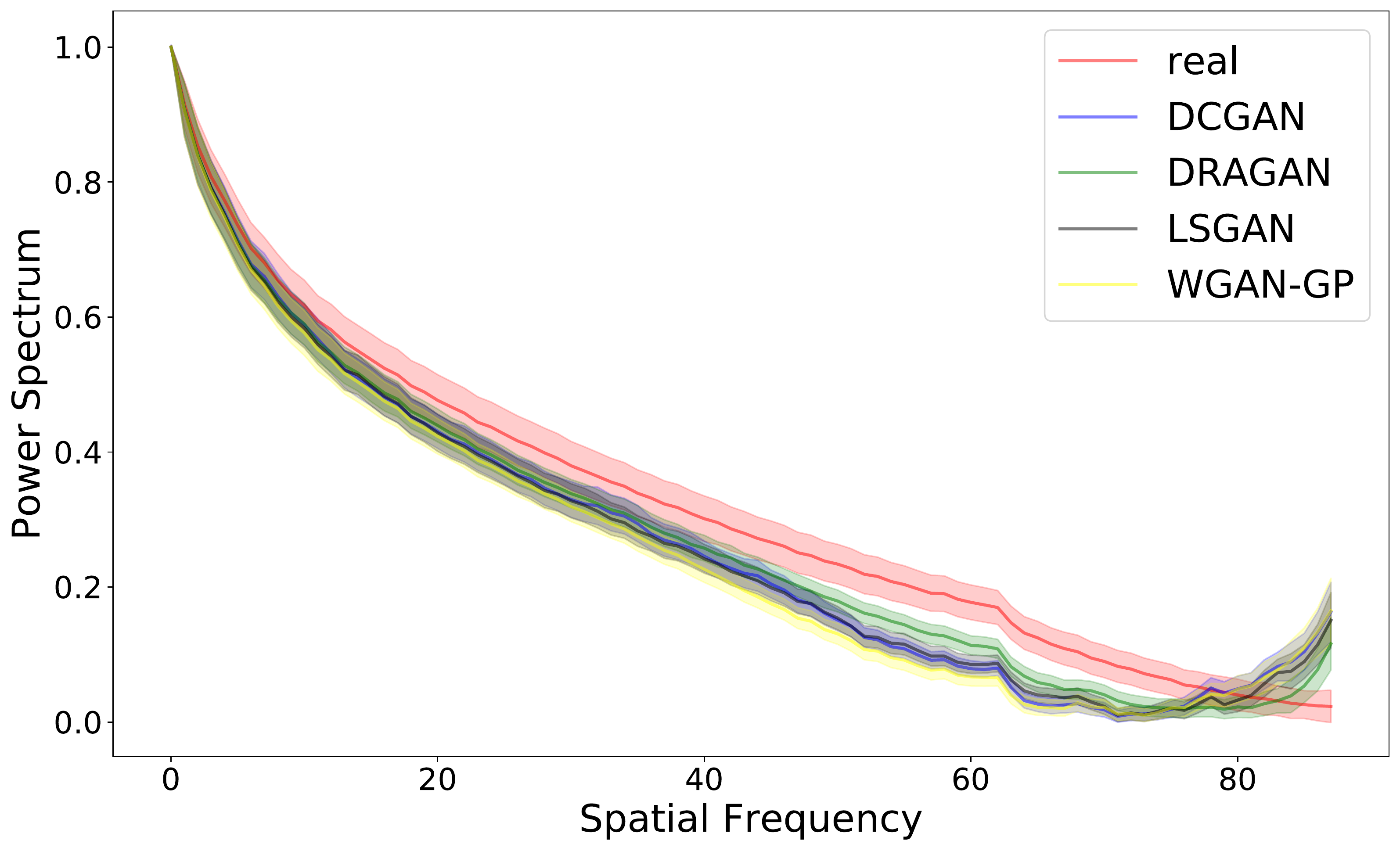}
   \includegraphics[width=0.9\linewidth]{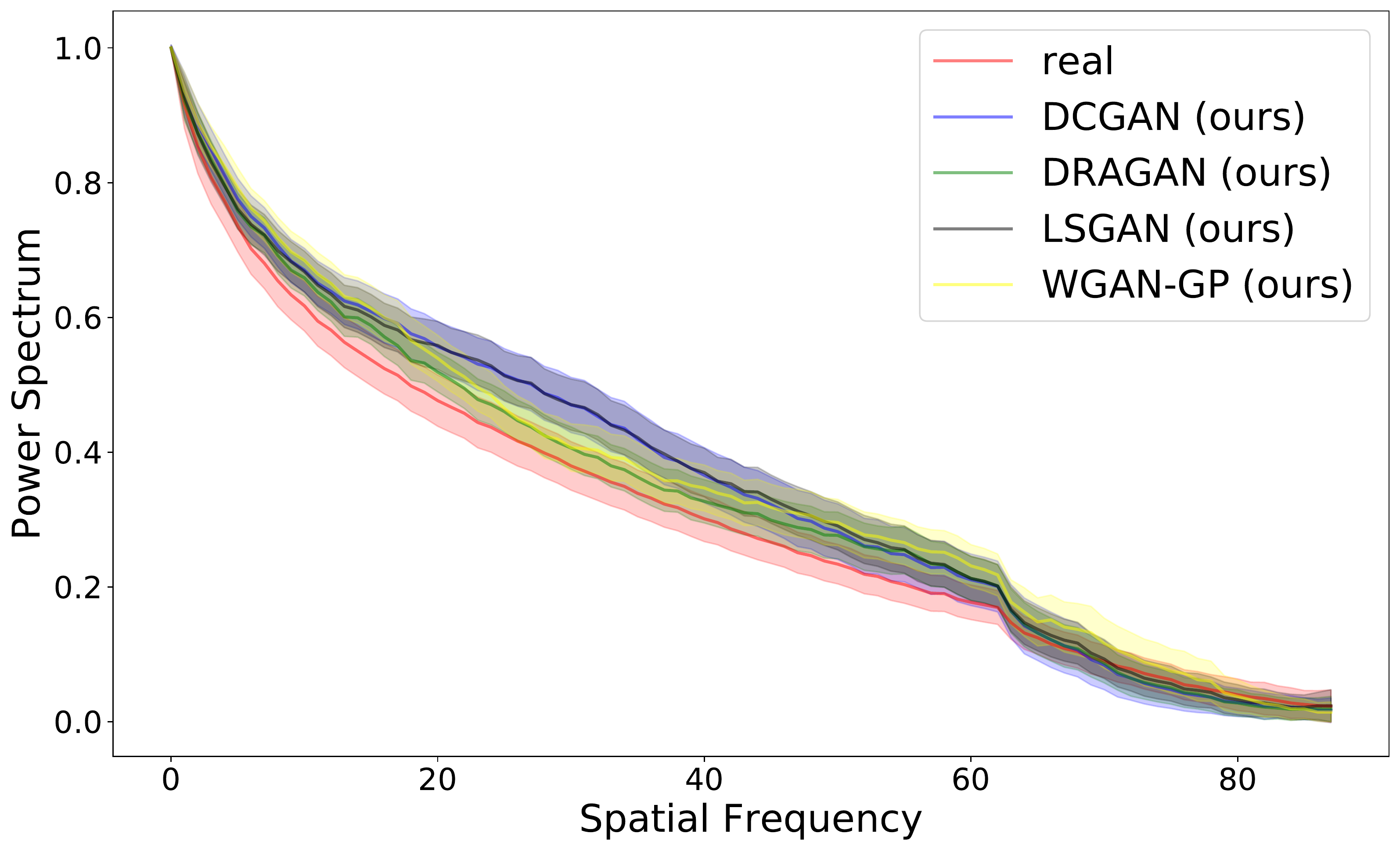}
   \caption{ Common up-convolution methods are inducing heavy spectral distortions into generated images. The \textbf{top} figure shows the statistics (mean and variance) after azimuthal integration over the power-spectrum (see Section \ref{sec:upconv:azi}) of real and GAN generated images. Evaluation on the \textit{CelebA} \cite{liu2015deep} data set, here all GANs (DCGAN \cite{radford2015unsupervised}, DRAGAN \cite{kodali2017convergence}, LSGAN \cite{mao2017least}, WGAN-GP \cite{gulrajani2017improved}) are using ``transposed convolutions'' (see Section \ref{sec:upconv:two}) for up-sampling.\\ \textbf{Bottom:} Results of the same experiments as above, adding our proposed spectral loss during GAN training.}
\label{fig:teaser}
\end{figure}
Generative convolutional deep neural networks have recently been used in a wide range of computer vision tasks: generation of photo-realistic images \cite{karras2017progressive,brock2018large}, image-to-image \cite{pathak2016context,iizuka2017globally,zhu2017toward,choi2018stargan,mo2018instanceaware,karras2019style} and text-to-image translations \cite{reed2016generative, dai2017towards, zhang2017stackgan, zhang2018stackgan++}, style transfer \cite{isola2017image,zhu2017unpaired,zhu2017toward,huang2018multimodal}, image inpainting \cite{pathak2016context,yeh2017semantic,li2017generative,iizuka2017globally,yu2018generative}, transfer learning \cite{bartunov2018few, clouatre2019figr,durall2019semi} or even for training semantic segmentation tasks \cite{luc2016semantic, xue2018segan}, just to name a few.

The most prominent generative neural network architectures are \textit{Generative Adversarial Networks (GAN)} \cite{goodfellow2014generative} and \textit{Variational Auto Encoders (VAE)} \cite{pu2016variational}. Both basic approaches try to approximate a latent-space model of the underlying (image) distributions from training data samples. Given such a latent-space model, one can draw new (artificial) samples and manipulate their semantic properties in various dimensions. While both GAN and VAE approaches have been published in many different variations, e.g. with different loss functions \cite{goodfellow2014generative, arjovsky2017wasserstein, gulrajani2017improved}, different latent space constraints \cite{mirza2014conditional,donahue2016adversarial, donahue2016adversarial, gurumurthy2017deligan,karras2019style} or various deep neural network (DNN) topologies for the generator networks \cite{radford2015unsupervised, nguyen2016synthesizing}, all of these methods have to follow a basic data generation principle: they have to transform samples from a low dimensional (often 1D) and low resolution latent space to the high resolution (2D image) output space. Hence, these generative neural networks must provide some sort of (learnable) up-scaling properties.

While all of these generative methods are steering the learning of their model parameters by optimization of some loss function, most commonly used losses are focusing exclusively on properties of the output image space, e.g. using convolutional neural networks (CNN) as discriminator networks for the implicit loss in an image generating GAN. This approach has been shown to be sufficient in order to generate visually sound outputs and is able to capture the data (image) distribution in image-space to some extent. However, it is well known that up-scaling operations notoriously alter the spectral properties of a signal \cite{jain1989fundamentals}, causing high frequency distortions in the output.

In this paper, we investigate the impact of up-sampling techniques commonly used in generator networks. The top plot of Figure \ref{fig:teaser} illustrates the results of our initial experiment, backing our working hypotheses that current generative networks fail to reproduce spectral distributions. Figure \ref{fig:teaser} also shows that this effect is independent of the actual generator network. 

\subsection{Related Work}

\subsubsection{Deepfake Detection}
We show the practical impact of our findings for the task of \textit{Deepfake detection}. The term \textit{deepfake} \cite{harris2018deepfakes, chesney2019deepfakes} describes the recent phenomenon of people misusing advances in artificial face generation via deep generative neural networks \cite{brundage2018malicious} to produce fake image content of celebrities and politicians. Due to the potential social impact of such fakes, \textit{deepfake detection} has become a vital research topic of its own.
Most approaches reported in the literature, like \cite{marra2018detection,afchar2018mesonet,cispa2965}, are themselves relying on CNNs and thus require large amounts of annotated training data. Likewise, \cite{hsu2018learning} introduces a deep
forgery discriminator with a contrastive loss function and
\cite{guera2018deepfake} incorporates temporal domain information by employing
Recurrent Neural Networks (RNNs) on top of CNNs.
\subsubsection{GAN Stabilization}
Regularizing GANs in order to facilitate a more stable training and to avoid mode collapse has recently drawn some attention. While \cite{unrolled} stabilize GAN training by unrolling the optimization of the discriminator,  \cite{stabgan} propose regularizations via noise as well as an efficient gradient-based approach. A stabilized GAN training based on octave convolutions has recently been proposed in \cite{durall2019stabilizing}. None of these approaches consider the frequency spectrum for regularization.
Yet, very recently, band limited CNNs have been proposed in \cite{dziedzic19a} for image classification with compressed models. In \cite{DBLP:journals/corr/abs-1906-08988}, first observations have been made that hint towards the importance of the power spectra on model robustness, again for image classification. In contrast, we propose to leverage observations on the GAN generated frequency spectra for training stabilization.

\subsection{Contributions}
The contributions of our work can be summarized as follows:
\begin{itemize}
\item We experimentally show the inability of current generative neural network architectures to correctly approximate the spectral distributions of training data.
\item We exploit these spectral distortions to propose a very simple but highly accurate detector for generated images and videos, i.e. a \textit{DeepFake} detector that reaches up to 100\% accuracy on public benchmarks. 
\item Our theoretical analysis and further experiments reveal that commonly used up-sampling units, i.e. up-convolutions, are causing the observed effects.
\item We propose a novel \textbf{spectral regularization} term which is able to compensate spectral distortions.
\item We also show experimentally that using spectral regularization in GAN training leads to more stable models and increases the visual output quality.   
\end{itemize}
The remainder of the paper is organized in as follows: Section \ref{sec:upconv} introduces common up-scaling methods and analyzes their negative effects on the spectral properties of images. In Section \ref{sec:correct}, we introduce a novel spectral-loss that allows to train generative networks that are able to compensate the up-scaling errors and generate correct spectral distributions. We evaluate our methods in Section \ref{sec:eval} using current architectures on public benchmarks.   

\section{The Spectral Effects of Up-Convolutions \label{sec:upconv} }

\subsection{Analyzing Spectral Distributions of Images using Azimuthal Integration over the DFT Power Spectrum \label{sec:upconv:azi}}
\begin{figure}[ht]
\centering
   \includegraphics[width=0.9\linewidth]{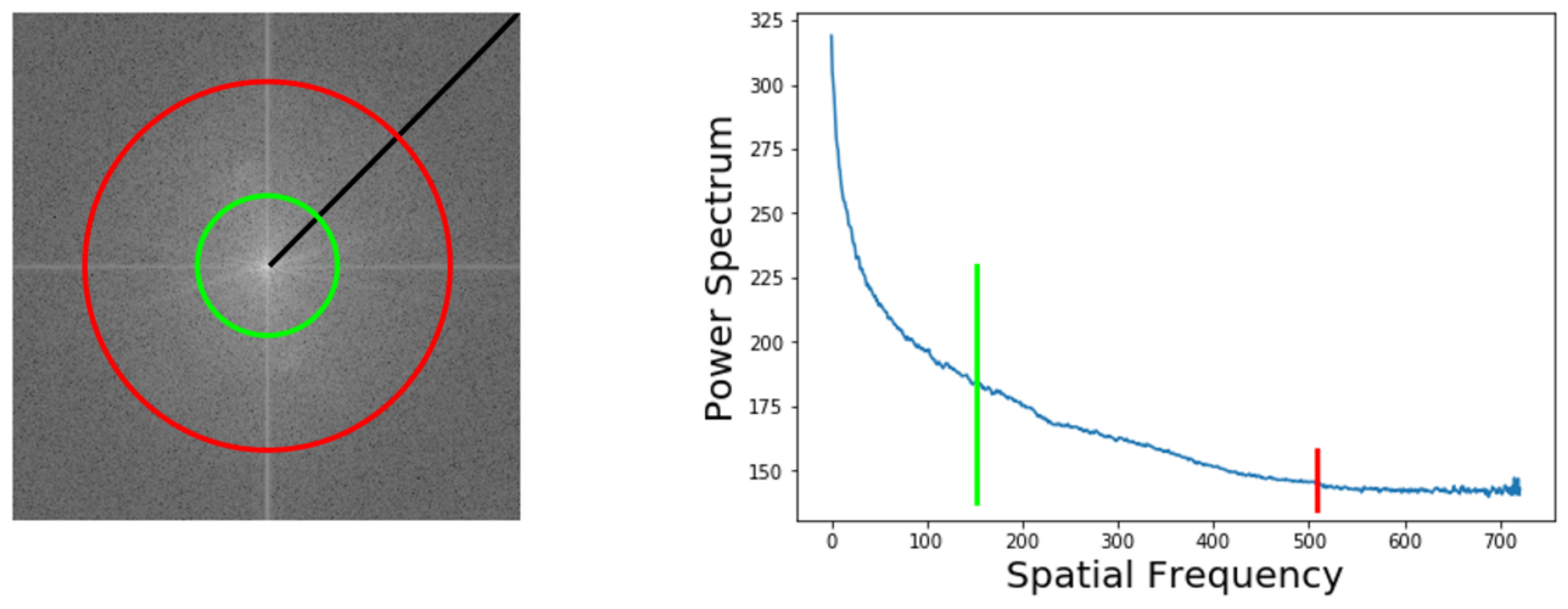}
   \caption{Example for the azimuthal integral (AI). (Left) 2D Power Spectrum of an image. (Right) 1D Power Spectrum: each frequency component is the radial integral over the 2D spectrum (red and green examples).}
\label{fig:azi}
\end{figure}

In order to analyze effects on spectral distributions, we rely on a simple but characteristic 1D representation of the Fourier power spectrum. We compute this spectral representation from the discrete Fourier Transform $\mathcal{F}$ of 2D (image) data $I$ of size $M\times N$,
\begin{align}
&{\mathcal{F}(I)}(k,\ell)=\sum_{m=0}^{M-1}\sum_{n=0}^{N-1}e^{-2\pi i\cdot\frac{jk}{M}}e^{-2\pi i\cdot\frac{j\ell}{N}}\cdot I(m,n), \\
&\mathrm{for }\quad k=0,\dots,M-1,\quad \ell=0,\dots,N-1,\nonumber
\end{align}
via azimuthal integration over radial frequencies $\phi$
\begin{align}\label{eq:AI}
&AI(\omega_k)=  \int_0^{2\pi} \|\mathcal{F}(I)\left(\omega_k\cdot \mathrm{cos}(\phi),\omega_k\cdot \mathrm{sin}(\phi)\right)\|^2 \mathrm{d}\phi\, \nonumber\\
&\mathrm{for }\quad k=0,\dots,M/2-1\, ,
\end{align}
assuming square images\footnote{$\rightarrow M=N$. We are aware that this notation is abusive, since $\mathcal{F}(I)$ is discrete. However, fully correct discrete notation would only over complicated a side aspect of our work. A discrete implementations of AI is provided on \url{https://github.com/cc-hpc-itwm/UpConv}. }.
Figure \ref{fig:azi} gives a schematic impression of this processing step.


\subsection{Up-convolutions in generative DNNs\label{sec:upconv:two} }
\begin{figure}[ht]
\centering
   \includegraphics[width=0.9\linewidth]{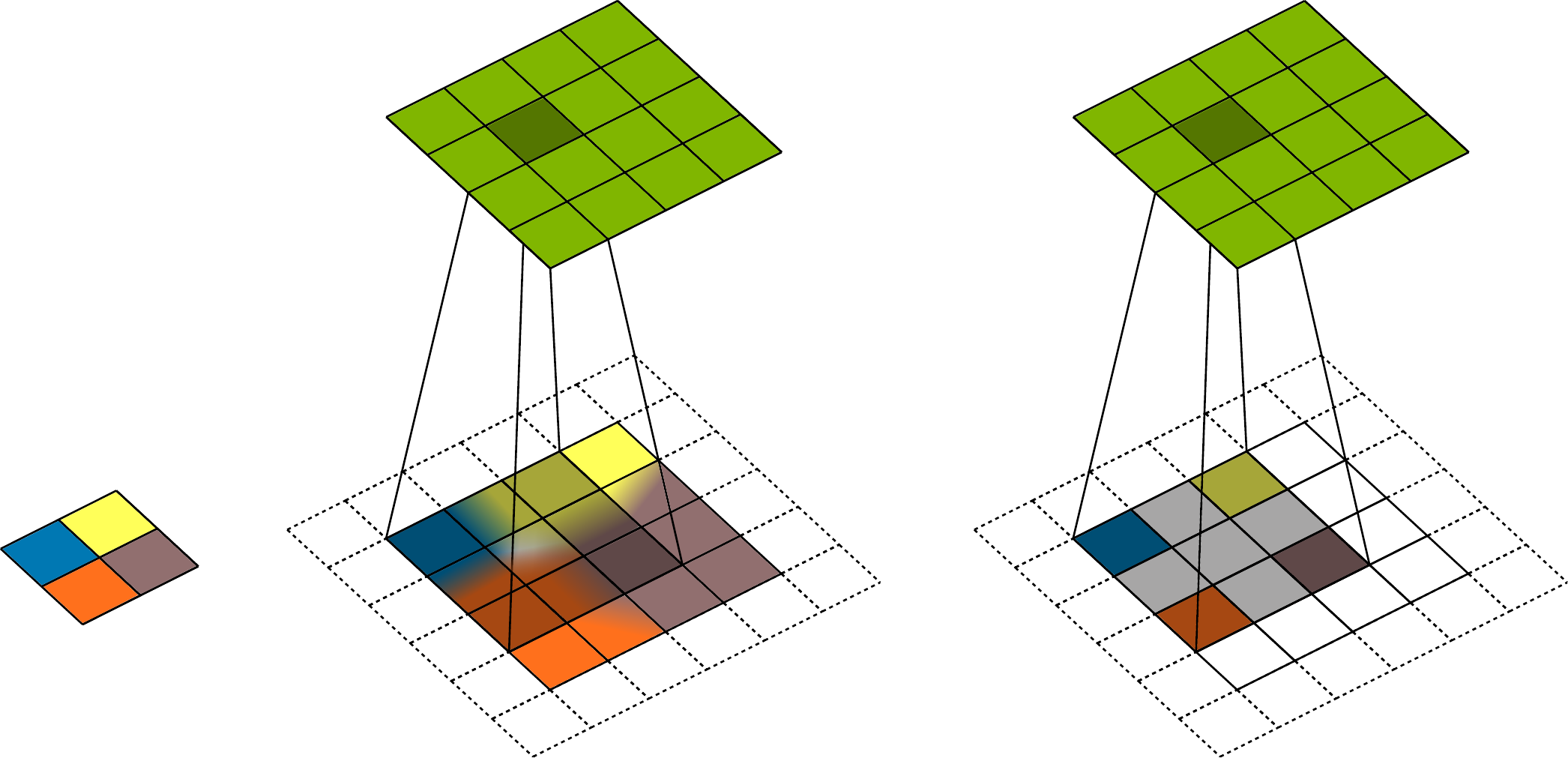}
   \caption{Schematic overview of the two most common up-convolution units. \textbf{Left: low resolution input image} (here $2\times 2$); \textbf{Center: up-convolution by interpolation} (up+conv) - the input is scaled via interpolation (bi-linear or nearest neighbor) and then convolved with a standard learnable filter kernel (of size $3\times 3$) to form the 5x5 output (green), \textbf{Right: transposed convolution} (transconv) - the input is padded with a ``bed of nails'' scheme (gray grid points are zero) and then convolved with a standard filter kernel to form the $5\times 5$ output (green).}
\label{fig:upconv:scheme}
\end{figure}

\begin{figure}[ht]
\centering
   \includegraphics[width=0.9\linewidth]{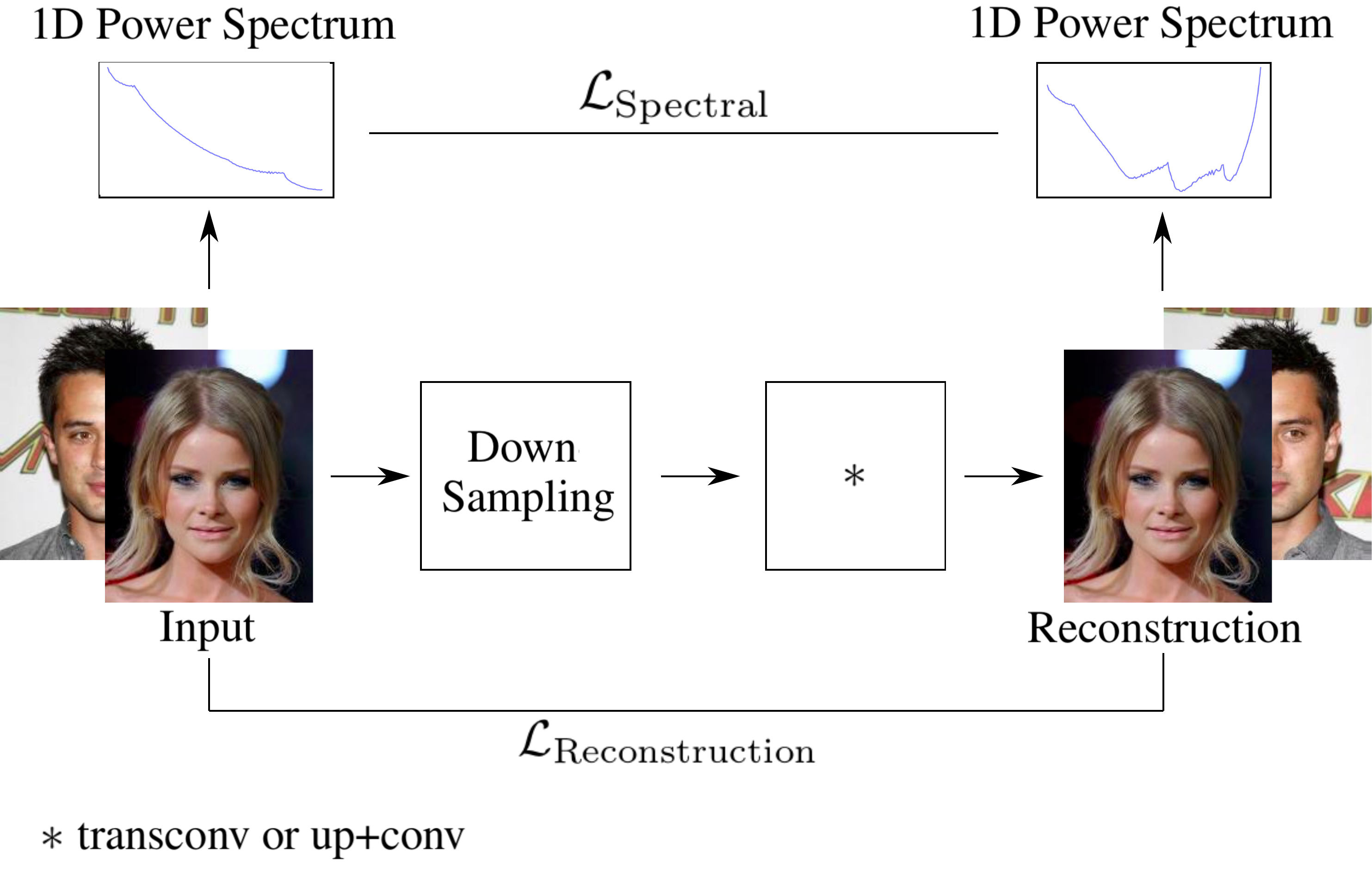}
   \caption{Schematic overview of the simple auto encoder (AE) setup used to demonstrate the effects of up-convolutions in Figure \ref{fig:upconv:res}, using only a standard MSE reconstruction loss (bottom) to train the AE on real images. We down-scale the input by a factor of two and then use the different up-convolution methods to reconstruct the original image size.  In Section \ref{sec:correct} we use the additional spectral loss (top) to compensate the spectral distortions (see Figure \ref{fig:correctedAE})  }.
\label{fig:upconv:AE}
\end{figure}

\begin{figure}[ht]
\centering
   \includegraphics[width=\linewidth]{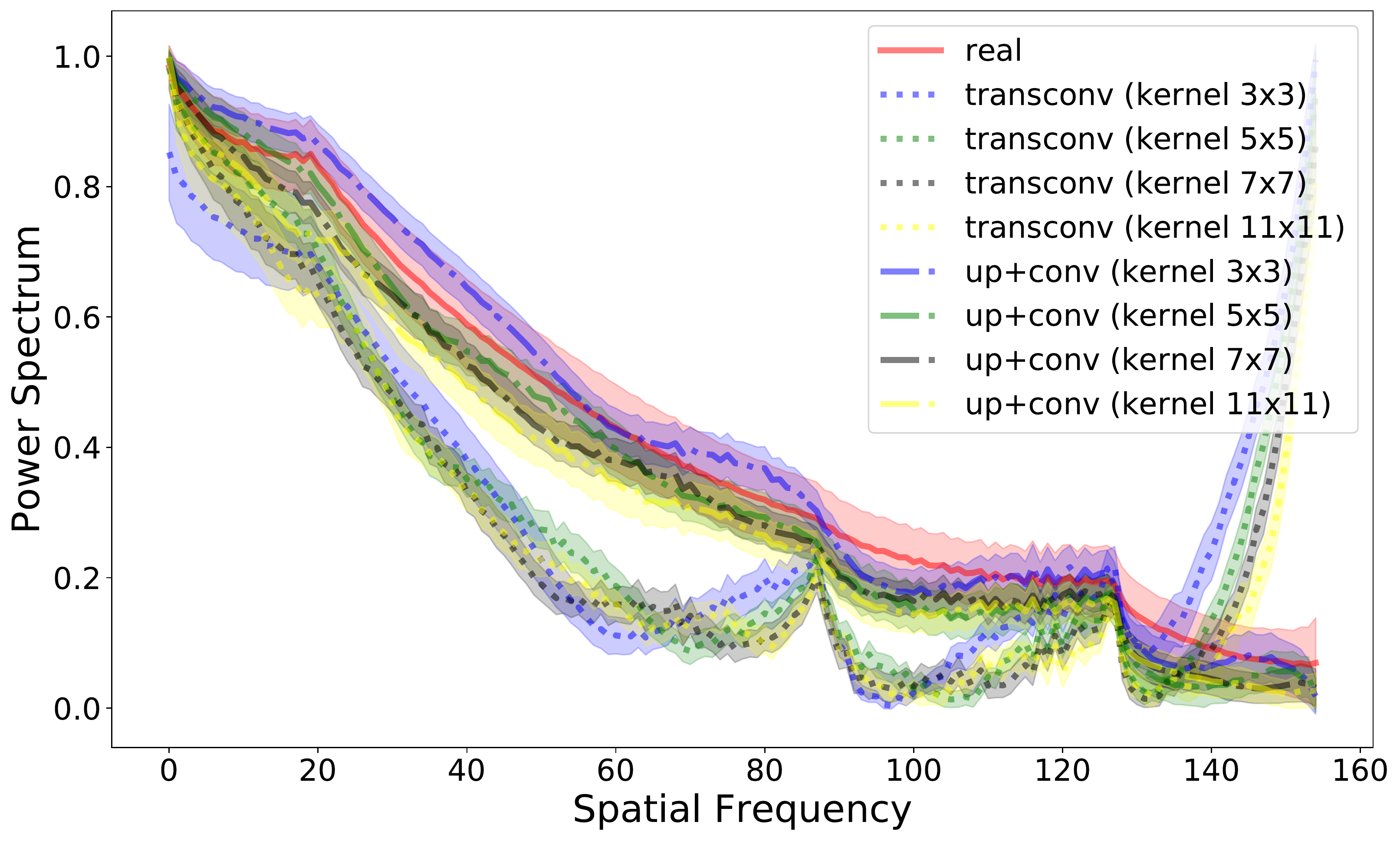}
   \caption{Effects of single up-convolution units (setup see Figure \ref{fig:upconv:AE}) on the frequency spectrum (azimuthal integral) of the output images. Both up-convolution methods have massive effects on the spectral distributions of the outputs. Transposed convolutions add large amounts high frequency noise while interpolation based methods (up+conv) are lacking high frequencies.}
\label{fig:upconv:res}
\end{figure}

\begin{figure}[ht]
\centering
   \includegraphics[width=0.9\linewidth]{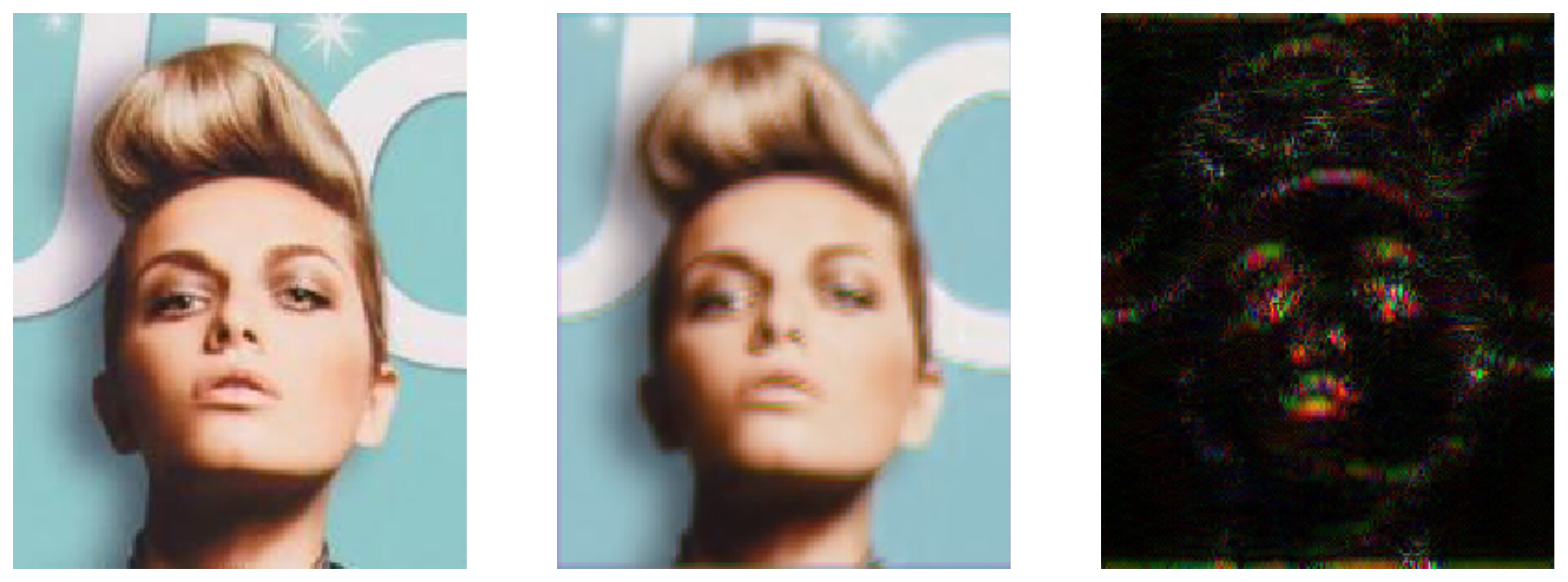}
   \includegraphics[width=0.9\linewidth]{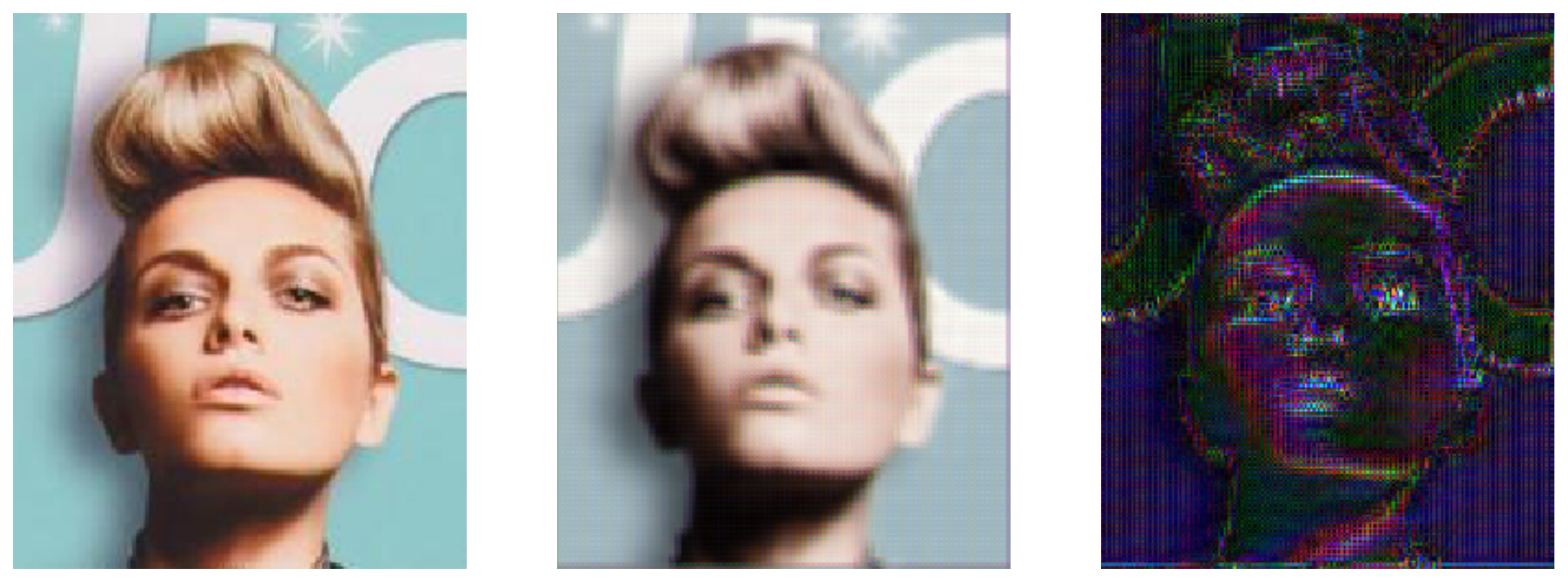}
\caption{Effects of spectral distortions on the image outputs in our simple AE setting. \textbf{Left:} original image; \textbf{Center:} AE output image; \textbf{Right:} filtered difference image . The \textbf{top row} shows the blurring effect of missing high frequencies in the (up+conv) case;  \textbf{Bottom row} shows the high frequency artifacts induces by (transconv).}
\label{fig:upconv:real-effects}
\end{figure}

Generative neural architectures like GANs produce high dimensional outputs, \eg images, from very low dimensional latent spaces. Hence, all of these approaches need to use some kind of up-scaling mechanism while propagating data through the network. The two most commonly used up-scaling techniques in literature and popular implementations frameworks (like TensorFlow \cite{tensorflow2015} and PyTorch \cite{paszke2017automatic}) are illustrated in Figure \ref{fig:upconv:scheme}: \textbf{up-convolution by interpolation} (up+conv) and \textbf{transposed convolution} (transconv) .\\
We use a very simple auto encoder (AE) setup (see Figure \ref{fig:upconv:AE}) for an initial investigation of the effects of up-convolution units on the spectral properties of 2d images after up-sampling. Figure \ref{fig:upconv:res} shows the different, but massive impact of both approaches on the frequency spectrum. Figure \ref{fig:upconv:real-effects} gives a qualitative result for a reconstructed image and shows that the mistakes in the frequency spectrum are relevant for the visual appearance.

\subsection{Theoretical Analysis\label{sec:theory}}
For the theoretic analysis, we consider, without loss of generality, the case of a one-dimensional signal $a$ and its discrete Fourier Transform $\hat{a}$
\begin{equation}
\hat{a}_k=\sum_{j=0}^{N-1}e^{-2\pi i\cdot\frac{jk}{N}}\cdot a_j, \quad \mathrm{for }\quad k=0,\dots,N-1.
\end{equation}
If we want to increase $a$'s spatial resolution by factor $2$
, we get 
\begin{align}
	\hat{a}^{up}_{\bar{k}}&=\sum_{j=0}^{2\cdot N-1}e^{-2\pi i\cdot\frac{j\bar{k}}{2\cdot N}}\cdot a^{up}_j\\
&=\sum_{j=0}^{N-1}e^{-2\pi i\cdot\frac{2\cdot j\bar{k}}{2\cdot N}}\cdot a_j
+ \sum_{j=0}^{N-1}e^{-2\pi i\cdot\frac{2\cdot (j+1)\bar{k}}{2\cdot N}}\cdot b_j,\\
&\mathrm{for}\quad \bar{k}=0,\dots,2N-1.\nonumber
\label{eq:theory2}
\end{align}
where $b_j=0$ for "bed of nails" interpolation (as used by \textit{transconv}) and $b_j=\frac{a_{j-1} + a_{j}}{2}$ for bi-linear interpolation (as used by \textit{up+conv}).

Let us first consider the case of $b_j=0$, \ie "bed of nails" interpolation. There, the second term in Eq. \eqref{eq:theory2} is zero. The first term is similar to the original Fourier Transform, yet with the parameter $k$ being replaced by $\bar{k}$. Thus, increasing the spatial resolution by a factors of $2$ leads to a scaling of the frequency axes by a factor of $\frac{1}{2}$. Let us now consider the effect from a sampling theory based viewpoint. 
It is 
\begin{align}
	\hat{a}^{up}_{\bar{k}}&=\sum_{j=0}^{2\cdot N-1}e^{-2\pi i\cdot\frac{j\bar{k}}{2\cdot N}}\cdot a^{up}_j\\
&=\sum_{j=0}^{2\cdot N-1}e^{-2\pi i\cdot\frac{j\bar{k}}{2\cdot N}}\cdot \sum_{t=-\infty}^{\infty}a^{up}_j\cdot\delta(j-2t)
	\label{eq:theory3}
\end{align}
since the point-wise multiplication with the Dirac impulse comb only removes values for which $a^{up}=0$. Assuming a periodic signal and applying the convolution theorem~\cite{conv-theorem}, we get
\begin{align}
\eqref{eq:theory3}
	&=\frac{1}{2}\cdot  \sum_{t=-\infty}^{\infty} \left(\sum_{j=-\infty}^{\infty}
e^{-2\pi i\cdot\frac{j\bar{k}}{2\cdot N}}a^{up}_j\right)\left(\bar{k}-\frac{t}{2}\right)\, ,
\end{align}
which equals to 
\begin{align}
	&\frac{1}{2}\cdot  \sum_{t=-\infty}^{\infty} \left(\sum_{j=-\infty}^{\infty}
e^{-2\pi i\cdot\frac{j\bar{k}}{N}}\cdot a_j\right)\left(\bar{k}-\frac{t}{2}\right)
\end{align}
by Eq. \eqref{eq:theory2}.
Thus, the "bed of nails upsampling" will create high frequency replica of the signal in $\hat{a}^{up}$. To remove these frequency replica, the upsampled signal needs to be smoothed appropriately. All observed spatial frequencies beyond $\frac{N}{2}$ are potential upsampling artifacts.
While it is obvious from a theoretical point of view, we also demonstrate practically in Figure \ref{fig:correctedAE-filter} that the correction of such a large frequency band is (assuming medium to high resolution images) is not possible with the commonly used $3\times 3$ convolutional filters.

In the case of bilinear interpolation, we have $b_j=\frac{a_{j-1} + a_{j}}{2}$ in Eq.  \eqref{eq:theory2}, which corresponds to an average filtering of the values of $a$ adjacent to $b_j$. This is equivalent to a point-wise multiplication of $a^{up}$ spectrum $\hat{a}^{up}$ with a sinc function by their duality and the convolution theorem, which suppresses artificial high frequencies. Yet, the resulting spectrum is expected to be overly low in the high frequency domain.


\section{Learning to Generate Correct Spectral Distributions \label{sec:correct}}

The experimental evaluations of our findings in the previous section and their application to detect generated content (see Section \ref{sec:deepfake}), raise the question if it would be possible to correct the spectral distortion induced by the up-convolution units used in generative networks. After all, usual network topologies contain learnable convolutional filters which follow the up-convolutions and potentially could correct such errors.      
\subsection{Spectral Regularization}
Since common generative network architectures are mostly exclusively using image-space based loss functions, it is not possible to capture and correct spectral distortions directly. Hence, we propose to add an additional spectral term to the generator loss: 
\begin{equation}\label{eq:spectral-loss}
	\mathcal{L}_{\mathrm{final}} = \mathcal{L}_{\mathrm{Generator}} + \lambda \cdot \mathcal{L}_{\mathrm{Spectral}}\, ,
\end{equation}
where $\lambda$ is the hyper-parameter that weights the influence of the spectral loss. Since we are already measuring spectral distortions using azimuthal integration $AI$ (see Eq. \eqref{eq:AI}), and $AI$ is differentiable, a simple choice for $\mathcal{L}_{\mathrm{Spectral}}$ is the \textbf{binary cross entropy} between the generated output $AI^{out}$ and the mean $AI^{real}$ obtained from real samples:

\begin{align}\label{eq:spectral-loss-real}
\mathcal{L}_{\mathrm{Spectral}} :=& -{1\over (M/2-1)} \sum_{i=0}^{M/2-1} AI^{real}_i \cdot \log(AI^{out}_i ) \nonumber \\
&+ (1-AI^{real}_i) \cdot  \log(1-AI^{out}_i )
\end{align}
Notice that $M$ is the image size and we use normalization by the $0^{th}$ coefficient ($AI_0$) in order to scale the values of the azimuthal integral to $[0,1]$.

\begin{figure}[h]
\centering
    \includegraphics[width=0.9\linewidth]{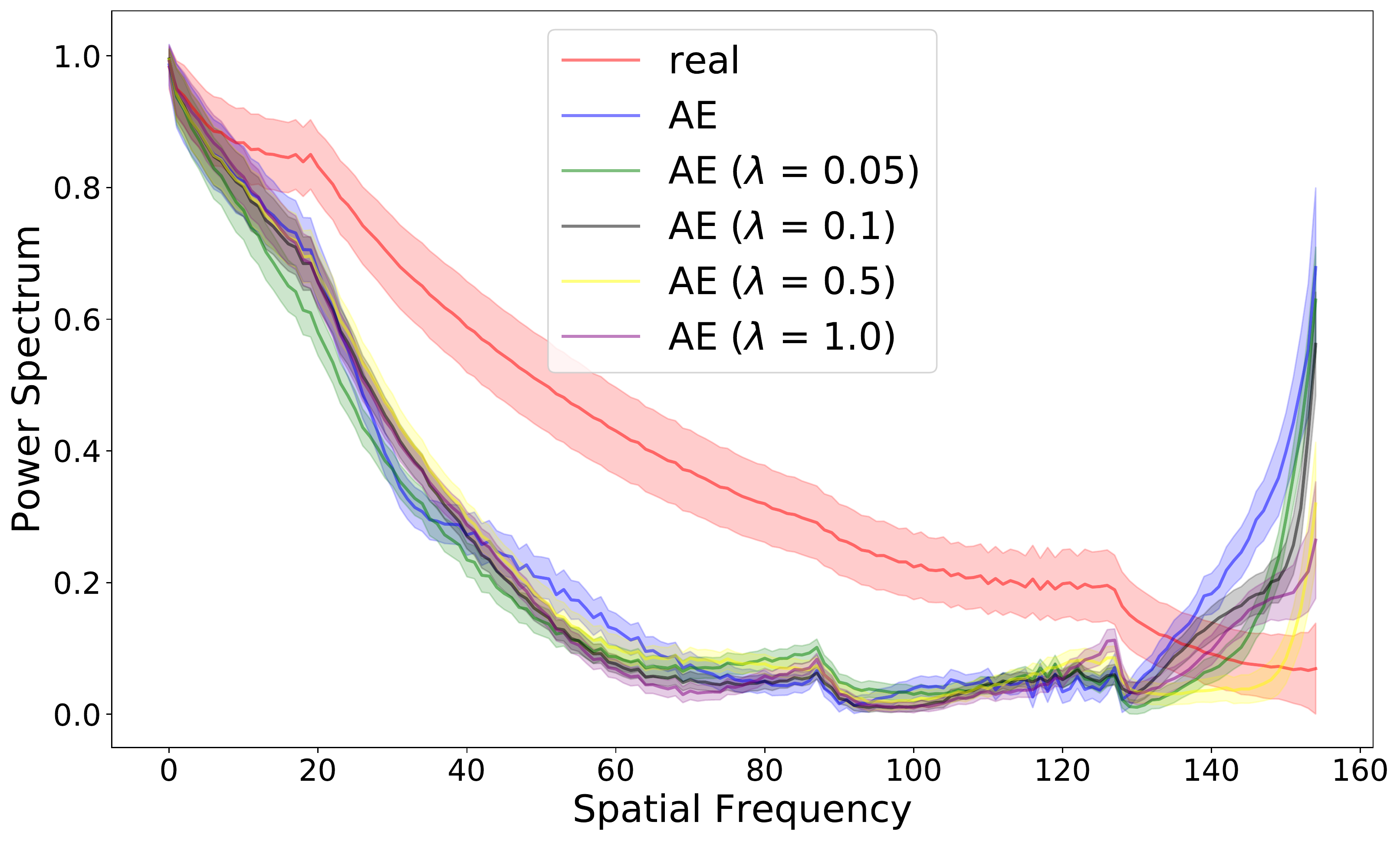}
    \caption{Auto encoder (AE) results with spectral loss by $\lambda$. Even if the spectral loss has a high weight, spectral distortions can not be corrected with a single $3\times 3$ convolutional layer. This result is in line with the findings from Section \ref{sec:theory}.}
\label{fig:correctedAE}
\end{figure}

The effects of adding our spectral loss to the AE setup from Section \ref{sec:upconv:two}  
for different values of $\lambda$  are shown in Figure \ref{fig:correctedAE}. As expected based on 
our theoretical analysis in sec. \ref{sec:theory},
the observed effects can not be corrected by a single, learned $3\times 3$ filter, even for large values $\lambda$. We thus need to reconsider the architecture parameters.



\subsection{Filter Sizes on Up-Convolutions}
\begin{figure}[h]
\centering
    \includegraphics[width=0.9\linewidth]{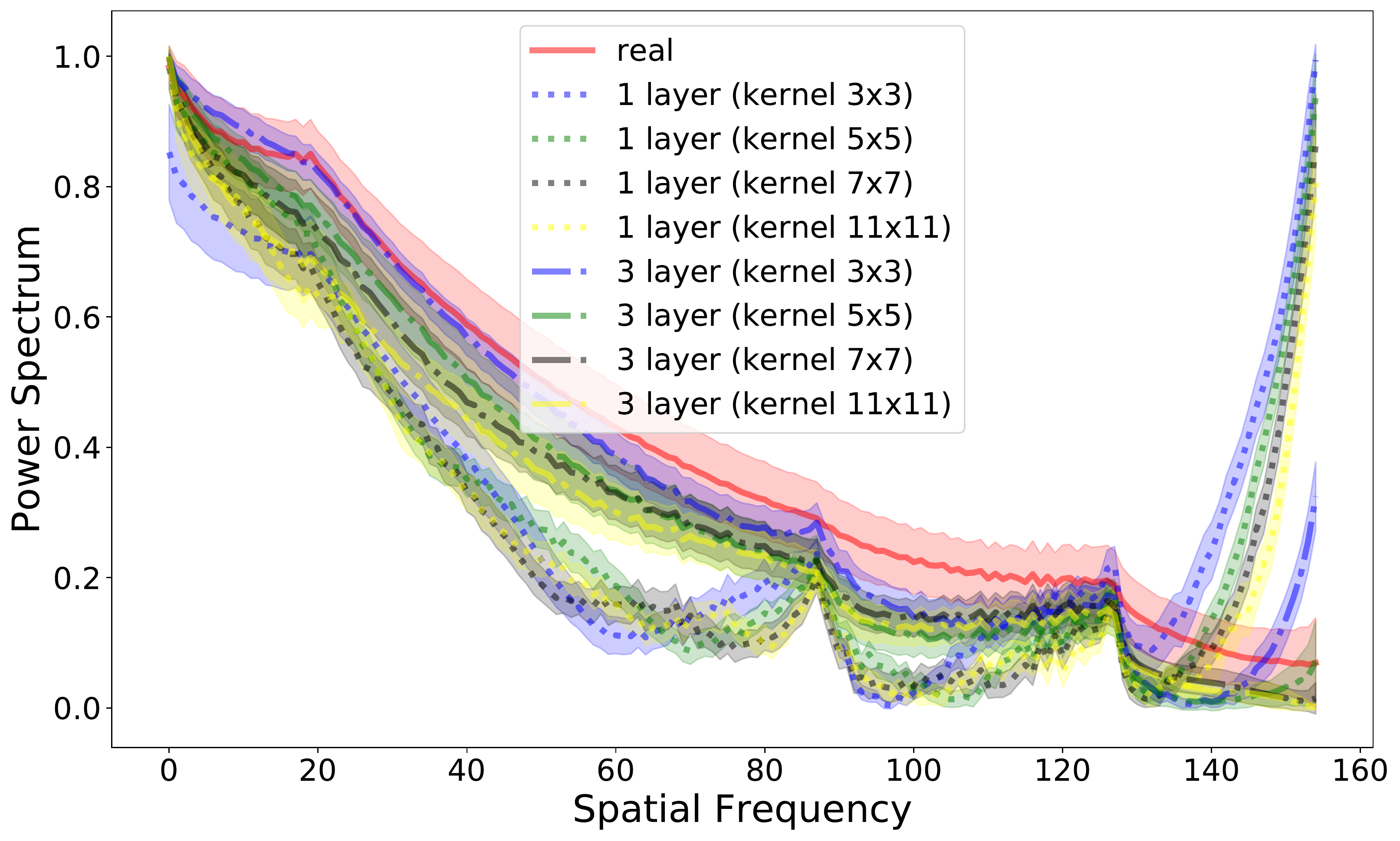}
    \caption{AE results with spectral loss by filter size of the convolution following the up-sampling step. The result heavily depends on the chosen filter size and number of convolutional layers.  With three $5\times 5$ convolutional filters available, the AE can greatly reduce spectral distortions using the proposed spectral loss.}
\label{fig:correctedAE-filter}
\end{figure}

In Figure \ref{fig:correctedAE-filter}, we evaluate our spectral loss on the AE from Section \ref{sec:upconv:two} with respect to filter sizes and the number of convolutional layers. 
We consider varying decoder filter sizes from $3\times 3$ to $11\times 11$ and 1 or 3 convolutional layers. While the spectral distortions from the up-sampling can not be removed with a single and even not with three $3\times 3$ convolutions, it can be corrected by the proposed loss when more, larger filters are learned. 

\section{Experimental Evaluation \label{sec:eval}}
We evaluate the findings of the previous sections in three different experiments, using prominent GAN architectures on public face generation datasets. Section \ref{sec:deepfake} shows that common face generation networks produce outputs with strong spectral distortions which can be used to detect artificial or ``fake'' images. In Section \ref{sec:use-spec-loss}, we show that our spectral loss is sufficient to compensate artifacts in the frequency domain of the same data. Finally, we empirically show in Section \ref{sec:pos-spec} that spectral regularization also has positive effects on the training stability of GANs.
\subsection{Deepfake Detection \label{sec:deepfake}}
\begin{figure*}[ht]
\centering
   \includegraphics[width=0.9\linewidth]{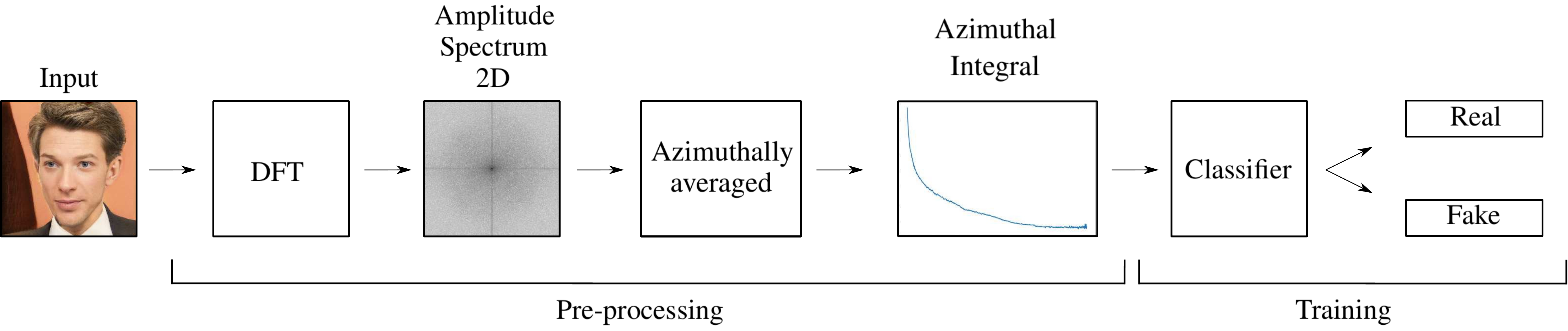}
   \caption{Overview of the processing pipeline of our approach. It contains two main blocks, a feature extraction block using DFT and a training block, where a classifier uses the new transformed features to determine whether the face is real or not. Notice that input images are converted to grey-scale before DFT.}
\label{fig:pipeline}
\end{figure*}
In this section, we show that the spectral distortions caused by the up-convolutions in state of the art GANs can be used to easily identify ``fake'' image data. Using only a small amount of annotated training data, or even an unsupervised setting, we are able to detect generated faces from public benchmarks with almost perfect accuracy. 
\subsubsection{Benchmarks}
We evaluate our approach on three different data sets of facial images, providing annotated data with different spacial resolutions:
\begin{itemize}
\item \textit{FaceForensics++} \cite{roessler2019faceforensicspp} contains a  DeepFake detection data set with 363 original video sequences of 28
paid actors in 16 different scenes, as well as over 3000 videos with face manipulations and their corresponding binary masks. All videos contain a
trackable, mostly frontal face without occlusions which enables automated
tampering methods to generate realistic forgeries. The resolution of the extracted face images varies, but is usually around $80\times 80\times 3$ pixels.   
\item The CelebFaces Attributes (\textit{CelebA}) dataset \cite{liu2015deep} consists of 202,599 celebrity face images with 40 variations in facial attributes. The dimensions of the face images are $178\times 218\times 3$, which can be considered to be a medium resolution in our context.
\item In order to evaluate high resolution $1024\times 1024\times 3$ images, we provide the new  \textit{Faces-HQ} \footnote{Faces-HQ data has a size of 19GB. Download: https://cutt.ly/6enDLYG. Also refer to \cite{durall2019unmasking}.} data set, which is a annotated collection of 40k publicly available images from \textit{CelebA-HQ} 
\cite{karras2017progressive}, \textit{Flickr-Faces-HQ} dataset \cite{karras2019style},
100K Faces project \cite{Faces}  and \textit{www.thispersondoesnotexist.com}.
\end{itemize}
\subsubsection{Method}

Figure \ref{fig:pipeline} illustrates our simple processing pipeline, extracting spectral features from samples via azimuthal integration (see Figure \ref{fig:azi}) and then using a basic SVM \cite{svm} classifier\footnote{SVM hyper-parameters can be found in the source code} for supervised and K-Means \cite{kmeans} for unsupervised fake detection. For each experiment, we randomly select training sets of different sizes and use the remaining data for testing. In order to handle input images of different sizes, we normalize the 1D power spectrum by the $0^{th}$ coefficient and scale the resulting 1D feature vector to a fixed size.

\subsubsection{Results}
Figure \ref{fig:1000} shows that real and ``fake'' faces form well delineated clusters in the high frequency range of our spectral feature space.  
The results of the experiments in Table \ref{tab:both} confirm that the distortions of the power spectrum, caused by the up-sampling units, are a common problem and allow an easy detection of generated content. This simple indicator even outperforms complex DNN based detection methods using large annotated training sets\footnote{Note: results of all other methods as reported by \cite{cispa2965}. The direct comparison of methods might be biased since \cite{cispa2965}  used the same real data but generated the fake data independently with different GANs.}.

\begin{figure}[!h]
\centering
   \includegraphics[width=0.9\linewidth]{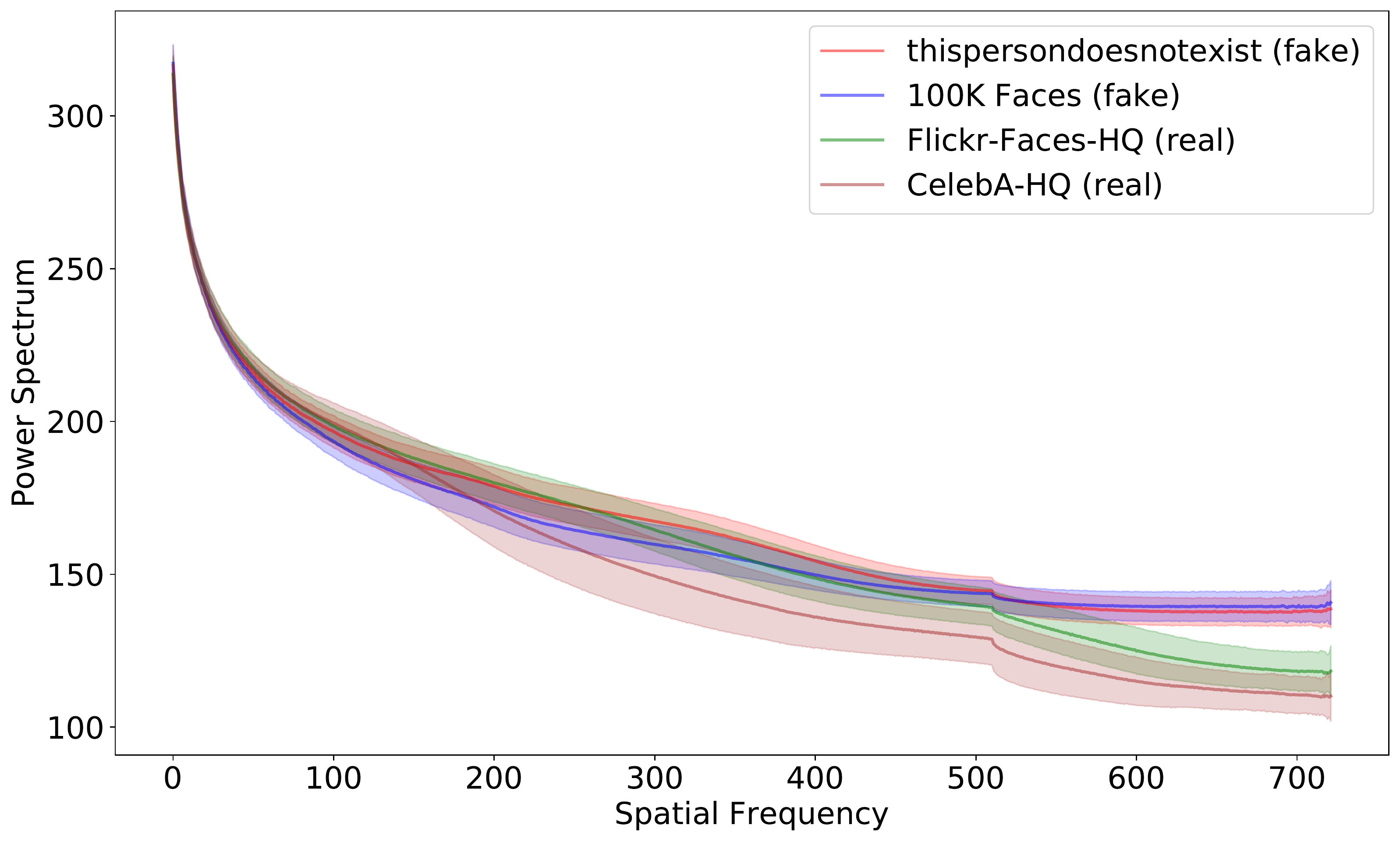}
   \caption{AI (1D power spectrum) statistics (mean and variance) of 1000 samples from each Faces-HQ sub-dataset. Clearly, real and ``fake'' images can be distinguished by their AI representation.}
\label{fig:1000}
\end{figure}

\begin{table}[ht]
\centering
\resizebox{0.45\textwidth}{!}{\begin{tabular}{c|cccc}
\hline
& \multicolumn{4}{c}{80\% (train) - 20\% (test)} \\
\cline{2-5}
data set & method & \# samples & supervised & unsupervised  \\
\hline
Faces-HQ & ours & 1000 & 100\% & \textbf{82\%}  \\
Faces-HQ & ours & 100 &  100\% & 81\%  \\
Faces-HQ & ours& 20 &  \textbf{100\%} & 75\%  \\
\hline
CelebA & ours & 2000 & \textbf{100\%} & \textbf{96\%}  \\
CelebA & \cite{cispa2965} & 100000 & 99.43\% & - \\
CelebA & \cite{fingerprint} & 100000 & 86.61\% & - \\
\hline
FaceForensics++& ours$^A$  & 2000 & 85\% & - \\
FaceForensics++& ours$^B$ & 2000&  \textbf{90\%} & -  \\
\hline
\end{tabular}}
\caption{Test accuracy. Our methods use SVM (supervised) and k-means (unsupervised) under different data settings. A) Evaluated on single frames. B) Accuracy on full video sequences via majority vote of single frame detections.}
\label{tab:both}
\end{table}

\begin{figure*}
\begin{subfigure}[t]{.25\linewidth}
\centering
\includegraphics[width=.95\linewidth]{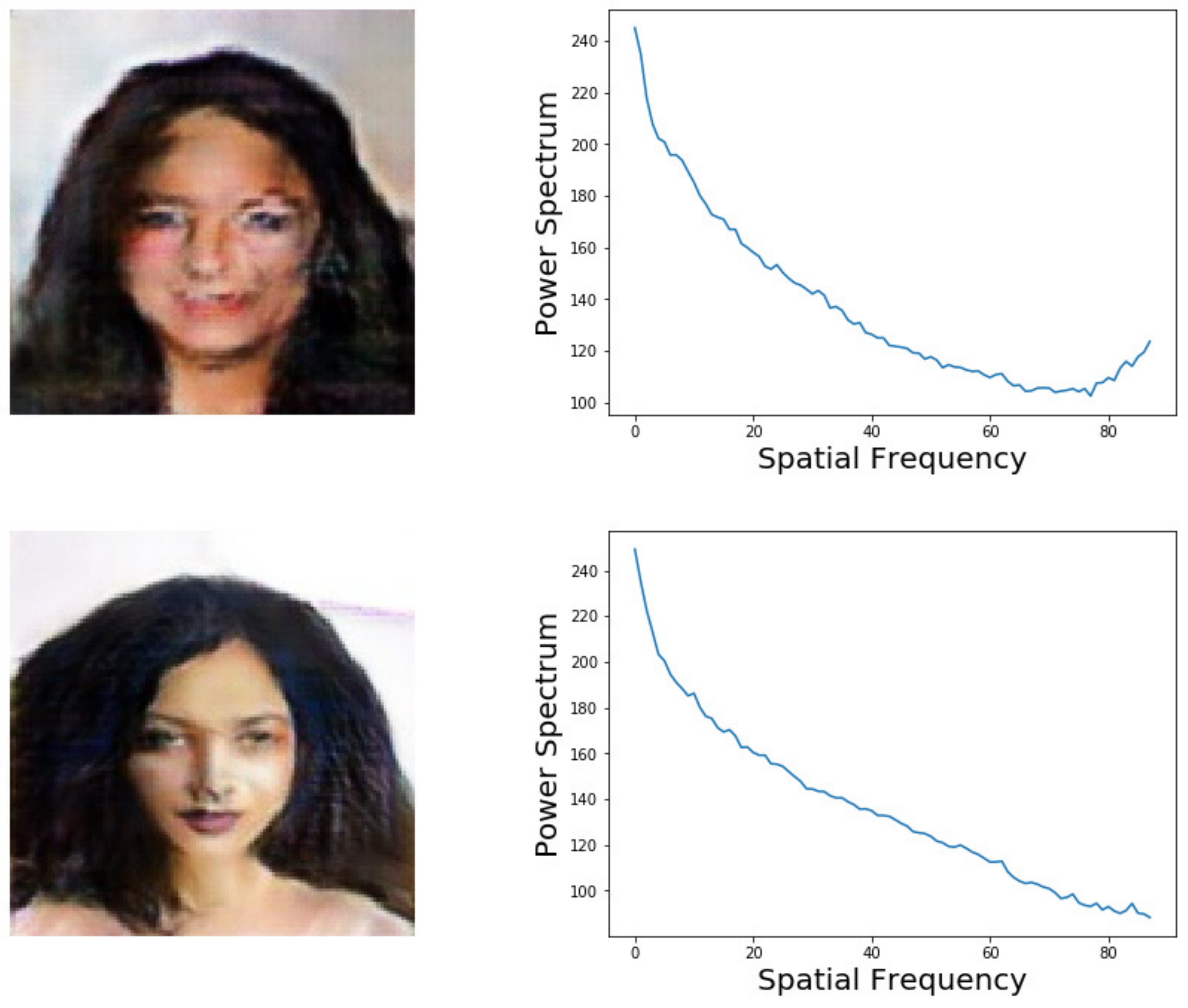}
\caption{DCGAN.}
\end{subfigure}%
\begin{subfigure}[t]{.25\linewidth}
\centering
\includegraphics[width=.95\linewidth]{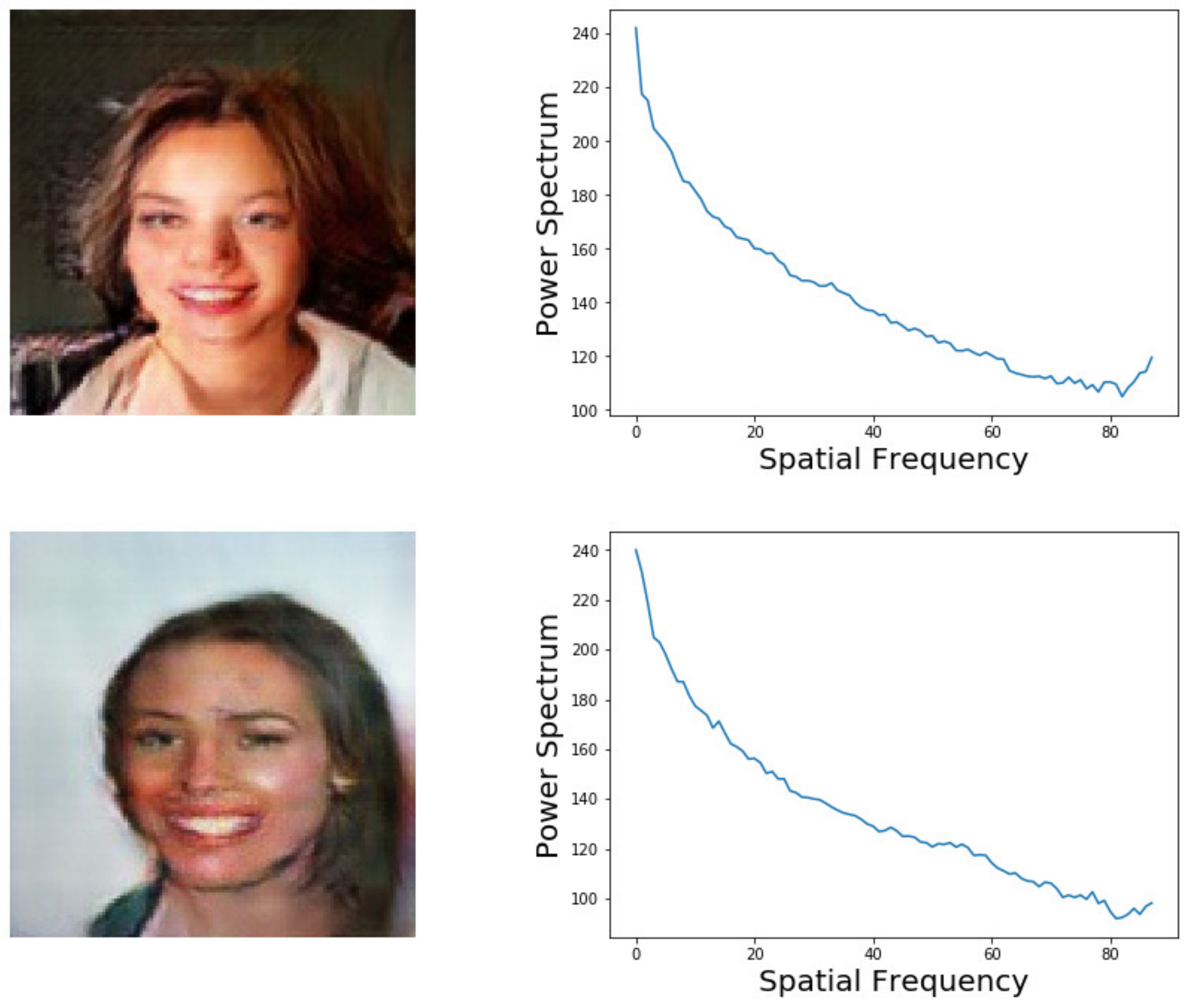}
\caption{DRAGAN.}
\end{subfigure}
\begin{subfigure}[t]{.25\linewidth}
\centering
\includegraphics[width=.95\linewidth]{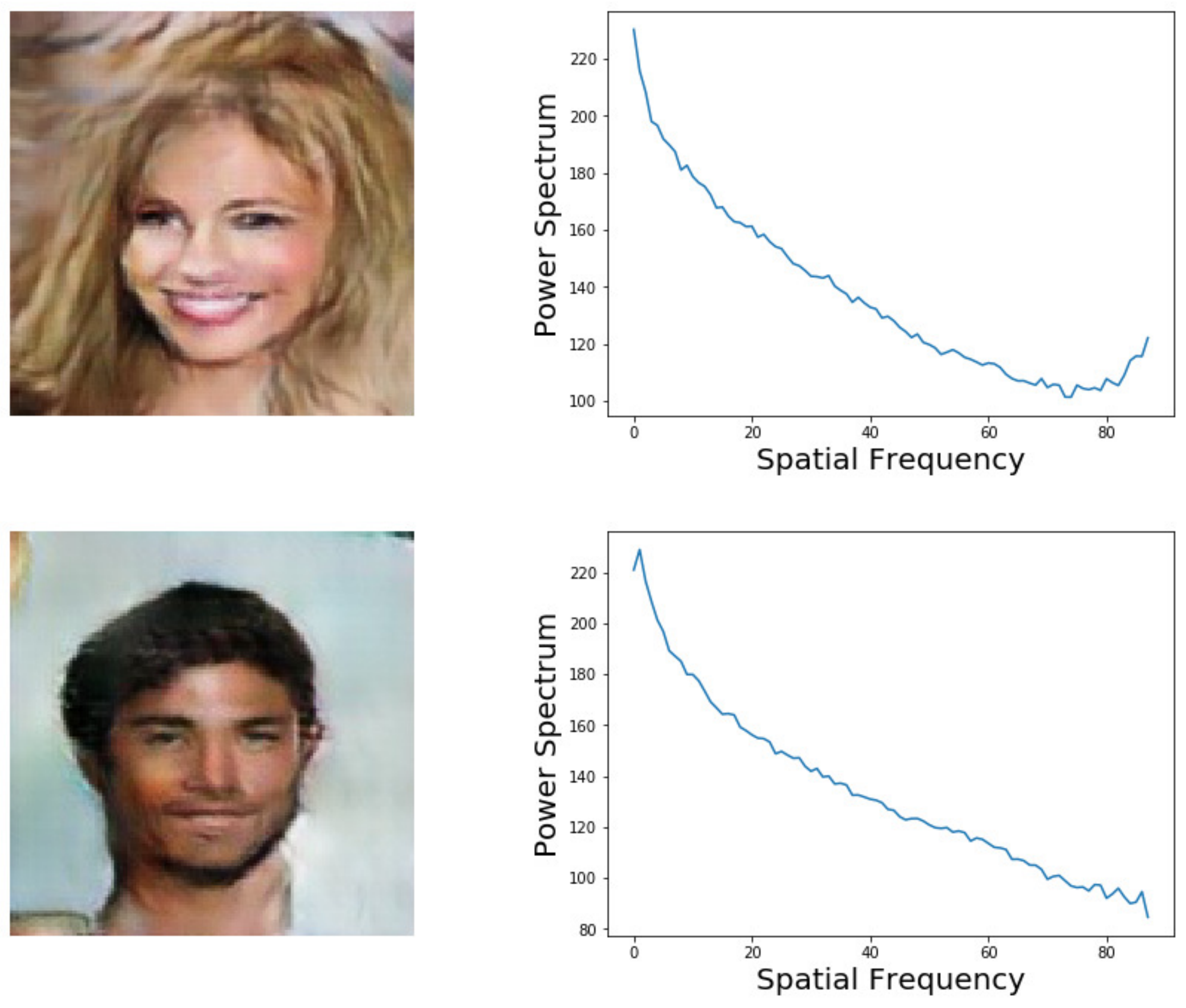}
\caption{LSGAN.}
\end{subfigure}%
\begin{subfigure}[t]{.25\linewidth}
\centering
\includegraphics[width=.95\linewidth]{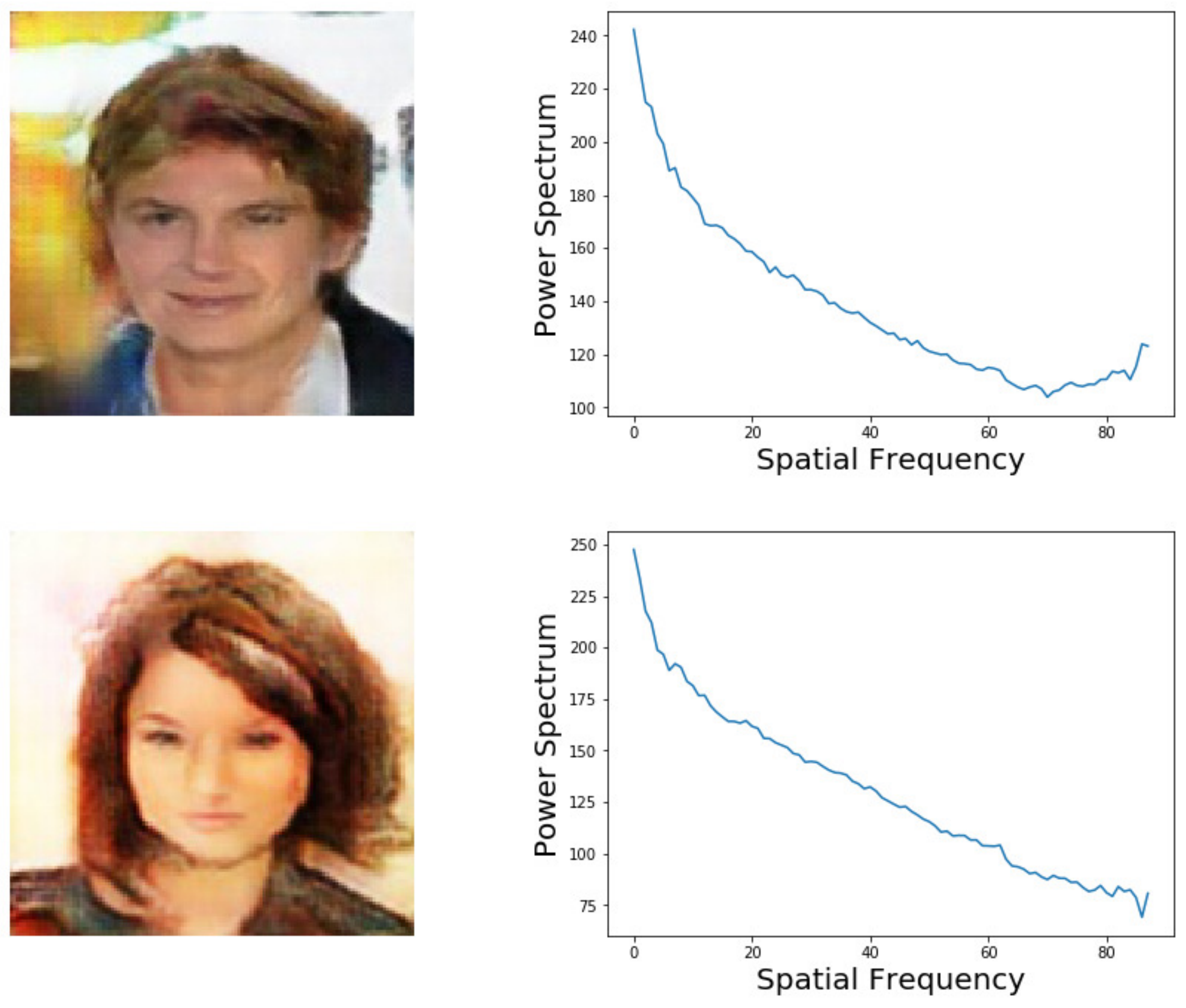}
\caption{WGAN.}
\end{subfigure}
\caption{Samples from the different types of GAN and their 1D Power Spectrum. \textbf{Top row:} samples produced by standard topologies. \textbf{Bottom row:} samples produced by standard topologies together with our spectral regularization technique.}
\label{fig:gans}
\end{figure*}
\begin{figure*}[ht]
\centering
   \includegraphics[width=0.9\linewidth]{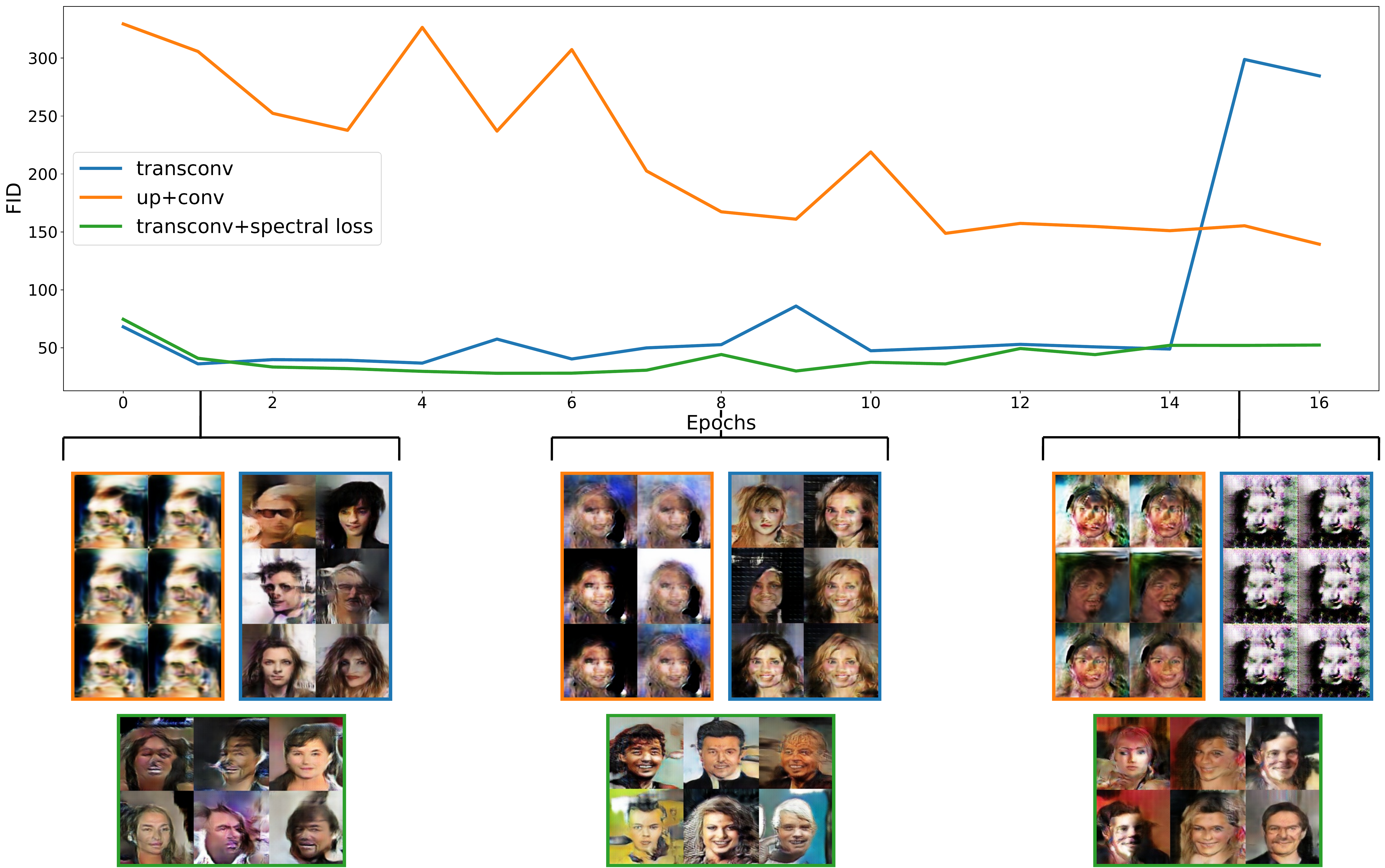}
   \caption{Correlation between FID values and GAN outputs for a DCGAN baseline on CelebA through out a training run. Low FID scores correspond to diverse but visually sound face image outputs. High FID scores indicate poor quality outputs and ``mode collapse'' scenarios where all generated images are bound to a very narrow sub-space of the original distribution. }
\label{fig:dcgan-collapse}
\end{figure*}
\subsection{Applying Spectral Regularization\label{sec:use-spec-loss}}
In this section, we evaluate the effectiveness of our regularization approach on the \textit{CelebA} benchmark, as in the experiment before.   
Based our theoretic analysis (see Section \ref{sec:theory}) and first AE experiments in Section \ref{sec:correct}, we extend existing GAN architectures in two ways: first, we add a spectral loss term (see Eq. \eqref{eq:spectral-loss-real}) to the generator loss. We use $1000$ unannotated real samples from the data set to estimate $AI^{real}$, which is needed for the computation of the spectral loss (see Eq. \eqref{eq:spectral-loss-real}). Second, we change the convolution layers after the last up-convolution unit to three filter layers with kernel size $5\times 5$. 
The bottom plot of Figure\ref{fig:teaser} shows the results for this experiment in direct comparison to the original GAN architectures. Several qualitative results produced without and with our proposed regularization are given in Figure \ref{fig:gans}.

\subsection{Positive Effects of Spectral Regularization\label{sec:pos-spec} }
By regularizing the spectrum, we achieve the direct benefit of producing synthetic images that not only look realistic, but also mimic the behaviour in the frequency domain. In this way, we are one step closer to sample images from the real distribution. Additionally, there is an interesting side-effect of this regularization. During our experiments, we noticed that GANs with a spectral loss term appear to be much more stable in terms of avoiding ``mode-collapse'' \cite{goodfellow2014generative} and better convergence. 
It is well known that GANs can suffer from challenging and unstable training procedures and there is little to no theory explaining this phenomenon. This makes it extremely hard to experiment with new generator variants, or to employ them in new domains, which drastically limits their applicability.

In order to investigate the impact of spectral regularization on the GAN training, we conduct a series of experiments. By employing a set of different baseline architectures, we assess the stability of our spectral regularization, providing quantitative results on the CelebA dataset. Our evaluation metric is the \textit{Fr\'echet Inception Distance} (FID) \cite{heusel2017gans}, which uses the Inception-v3 \cite{szegedy2016rethinking} network pre-trained on ImageNet \cite{deng2009imagenet} to extract features from an intermediate layer.

\begin{figure}[h]
\centering
   \includegraphics[width=0.9\linewidth]{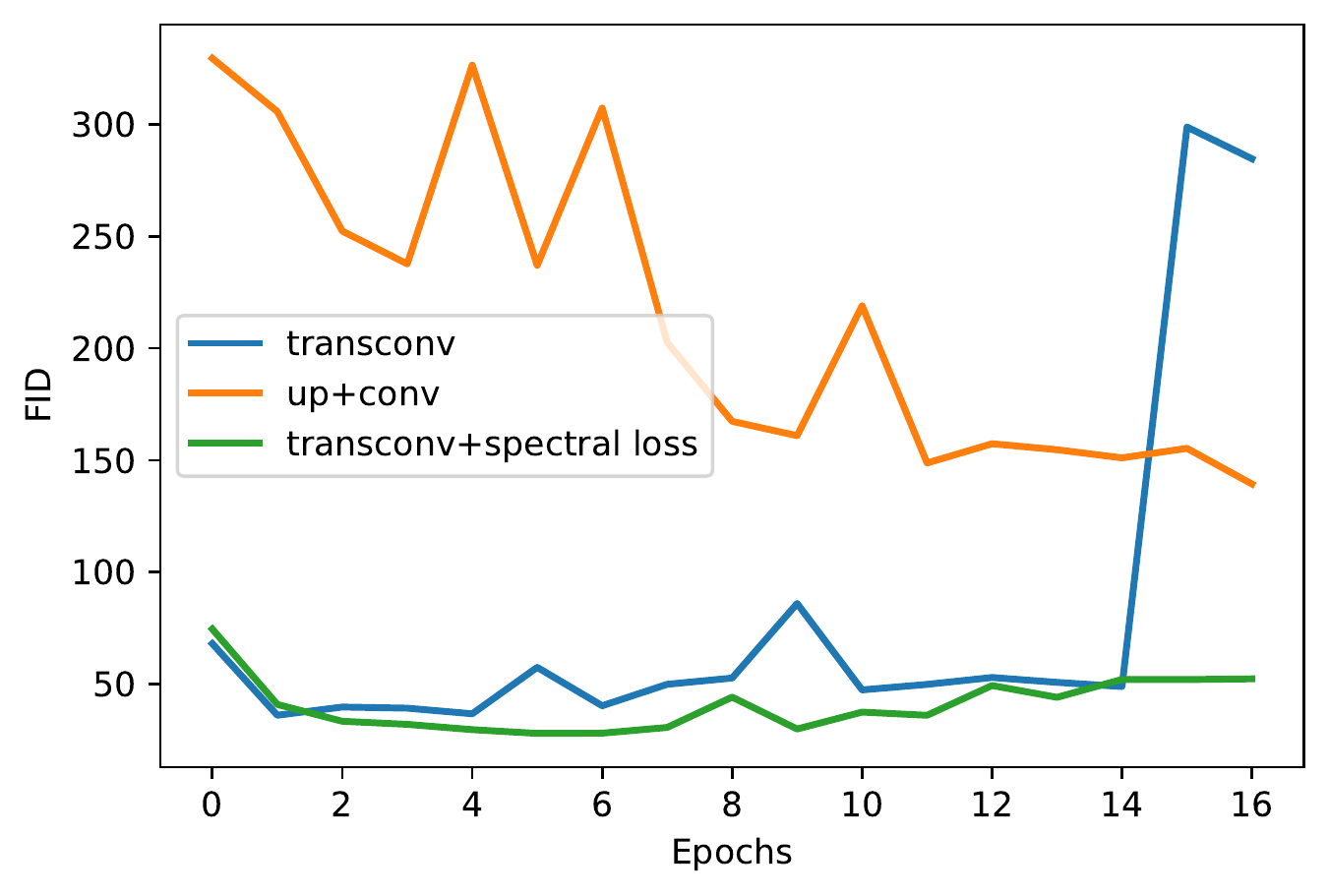}
   \caption{FID (lower is better) over training time for DCGAN baselines with and without spectral loss (here $\lambda = 2$).  While the \textit{up+conv} variant of DCGAN is failing to improve, the FID score over the training time in the  \textit{transconv} version is converging but unstable. Only our spectral loss variant is able to achieve low and stable FID scores.}
\label{fig:dcgan}
\end{figure}

\begin{figure}[h]
\centering
    \includegraphics[width=0.9\linewidth]{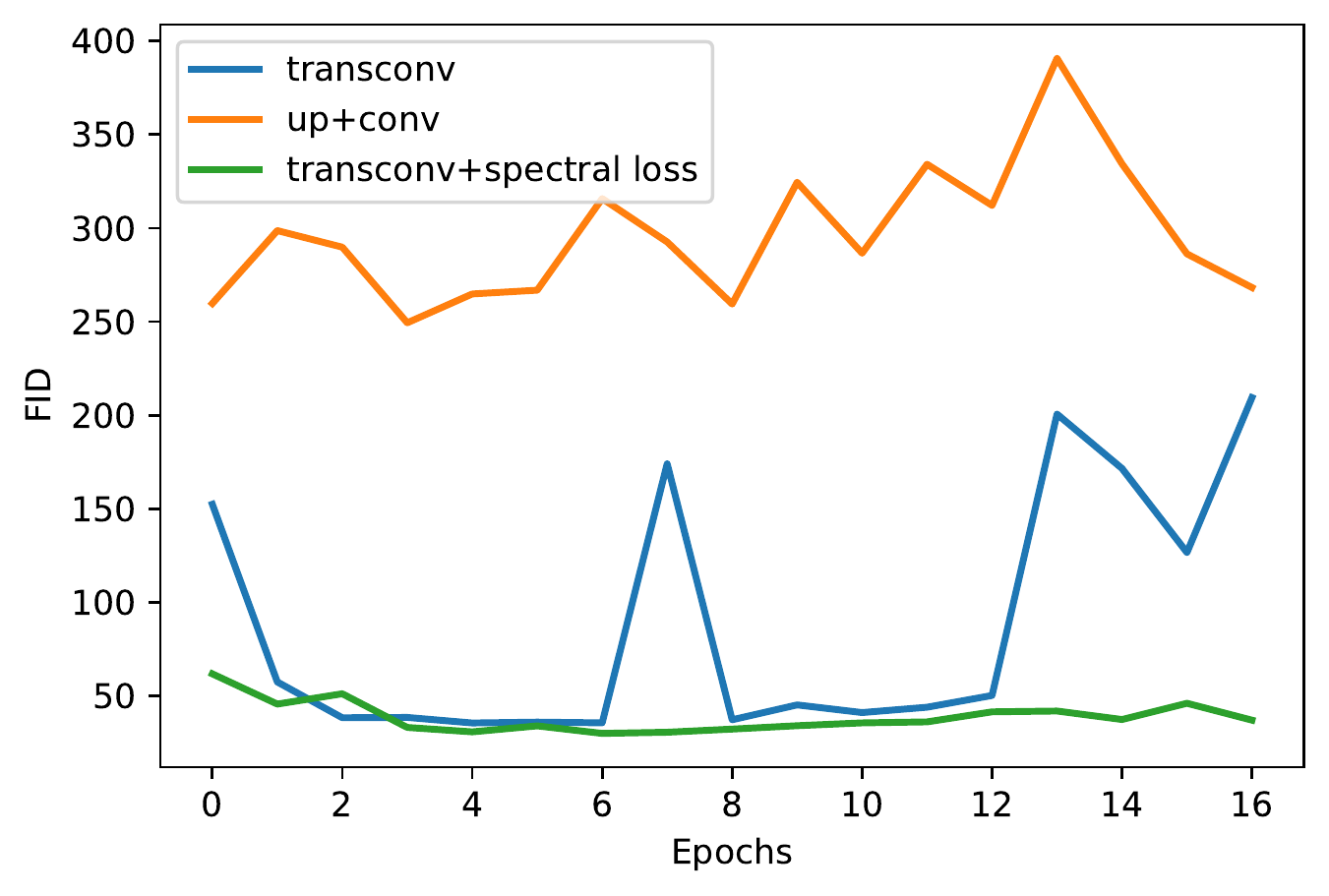}
    \caption{FID (lower is better) over training time for LSGAN baselines with and without spectral loss (here $\lambda = 0.5$). As for DCGANS, is the \textit{up+conv} variant of LSGAN failing to improve the FID score over the training time. The  \textit{transconv} version is converging but unstable. Again, only our spectral loss variant is able to achieve low and stable FID scores. }
\label{fig:lsgan}
\end{figure}

Figures \ref{fig:dcgan} and \ref{fig:lsgan} show the FID evolution along the training epochs, using a baseline GAN implementation with different up-convolution units and a corresponding version with spectral loss. These results show an obvious positive effect in terms of the FID measure, where spectral regularization k.pdf a stable and low FID through out the training while unregularized GANs tend to ``collapse''. Figure \ref{fig:dcgan-collapse} visualizes the correlation between high FID values and failing GAN image generations.

\section{Discussion and Conclusion}
We showed that common ``state of the art'' convolutional generative networks, like popular GAN image generators fail to approximate the spectral distributions of real data. This finding has strong practical implications: not only can this be used to easily identify generated samples, it also implies that all approaches towards training data generation or transfer learning are fundamentally flawed and it can not be expected that current methods will be able to approximate real data distributions correctly. However, we showed that there are simple methods to fix this problem: by adding our proposed spectral regularization to the generator loss function and increasing the filter sizes of the final generator convolutions to at least $5 \times 5$, we were able to compensate the spectral errors.
Experimentally, we have found strong indications that the spectral regularization has a very positive effect on the training stability of GANs. While this phenomenon needs further theoretical investigation, intuitively this makes sense as it is known that high frequent noise can have strong effects on CNN based discriminator networks, which might cause overfitting of the generator.\\   

\noindent Source code available:\\ \url{https://github.com/cc-hpc-itwm/UpConv}

\newpage
{\small
\bibliographystyle{ieee}
\bibliography{egbib}
}

\newpage
\section*{Supplemental Material}

The supplementary material of our paper contains additional details on the presented experiments, as well as some support experiments that might help to get a better understanding of the spectral properties of up-convolution units.   

\section{Using Spectral Distortions to Detect Deepfakes}
In this section, we provide more detailed results of the experiments presented in section 4.1 of the paper.

\subsection{More Details on the used Datasets}
\subsubsection{Faces-HQ}
To the best of our knowledge, currently no public dataset is providing high resolution images with annotated fake and real faces. Therefore, we have created our own data set from established sources, called \textit{Faces-HQ}\footnote{Faces-HQ data has a size of 19GB. Download: https://cutt.ly/6enDLYG}. In order to have a sufficient variety of faces, we have chosen to download and label the images available from the \textit{CelebA-HQ} data set
\cite{karras2017progressive}, \textit{Flickr-Faces-HQ} data set \cite{karras2019style},
100K Faces project \cite{Faces}  and \textit{www.thispersondoesnotexist.com}. In total, we have
collected 40K high quality images, half of them real and the other half
fake faces. Table
\ref{tab:summary} contains a summary.\\
\noindent\textbf{Training Setting:}  we divide  the transformed data into
training and testing sets, with  20\% for the testing stage and use the
remaining 80\% as the training set. Then, we train a classifier with the
training data and finally evaluate the accuracy on the testing set.
\begin{table}[h]
\centering
\resizebox{0.48\textwidth}{!}{\begin{tabular}{c|ccc}
\hline
& \# of samples & category & label \\
\hline
CelebA-HQ data set \cite{karras2017progressive} & 10000 & Real & 0 \\
Flickr-Faces-HQ data set \cite{karras2019style} & 10000 & Real & 0 \\
100K Faces project \cite{Faces} & 10000 & Fake & 1 \\
www.thispersondoesnotexist.com & 10000 & Fake & 1 \\
\hline
\end{tabular}}
	\caption{\textit{Faces-HQ} data set structure.}
\label{tab:summary}
\end{table}

\subsubsection{CelebA}
The CelebFaces Attributes (\textit{CelebA}) dataset \cite{liu2015deep} consists of 202,599 celebrity face images with 40 variations in facial attributes. The dimensions of the face images are 178x218x3, which can be considered to be a medium-resolution in our context.\\
\noindent\textbf{Training Setting:}
While we can use the real images from the \textit{CelebA} dataset directly, we need to generate the fake examples on our own.Therefore we use the real dataset to train one DCGAN  \cite{radford2015unsupervised}, one DRAGAN \cite{kodali2017convergence}, one LSGAN \cite{mao2017least} and one WGAN-GP \cite{gulrajani2017improved}  to
generate realistic fake images. We
split the dataset into 162,770 images for training and 39,829 for testing, and
we crop and resize the initial 178x218x3 size images to 128x128x3. Once the
model is trained, we can conduct the classification experiments on
medium-resolution scale.

\subsubsection{FaceForensics++}
\textit{FaceForensics++} \cite{roessler2019faceforensicspp} is a collection of image forensic datasets, containing video sequences that have been modified with different automated
face manipulation methods. One subset is the DeepFakeDetection
Dataset, which contains 363 original sequences from 28
paid actors in 16 different scenes as well as over 3000 manipulated videos
using DeepFakes and their corresponding binary masks. All videos contain a
trackable, mostly frontal face without occlusions which enables automated
tampering methods to generate realistic forgeries.\\
\noindent\textbf{Training Setting:}
the employed pipeline for this dataset is the same as for \textit{Faces-HQ} dataset and \textit{CelebA}, but with an additional block. Since the DeepFakeDetection dataset contains videos,
we first need to extract the frame and then crop the inner faces from them. Due
to the different content of the scenes of the videos, these cropped faces have
different sizes. Therefore, we interpolate the 1D Power Spectrum to a fix
size (300) and normalizes it dividing it by the $0^{th}$ frequency component.

\subsection{Experimental Results}

\subsubsection{Spectral Distributions}
The following figures \ref{fig:1000}, \ref{fig:1000deepfak} and  \ref{fig:1000gan} show the spectral (AI) distributions of all datasets. In all three cases, it is evident that a classifier should be able to separate real and fake samples. Also, based on our theoretical analysis (see section 2.3 in the paper), one can assume that the generators in used Face-HQ and FaceForensics++ datasets used \textit{up+conv} based up-convolutions or successively blurred the generated images (due to the drop in high frequencies). CelebA based fakes used \textit{transconv}.

\begin{figure}[!h]
\centering
   \includegraphics[width=\linewidth]{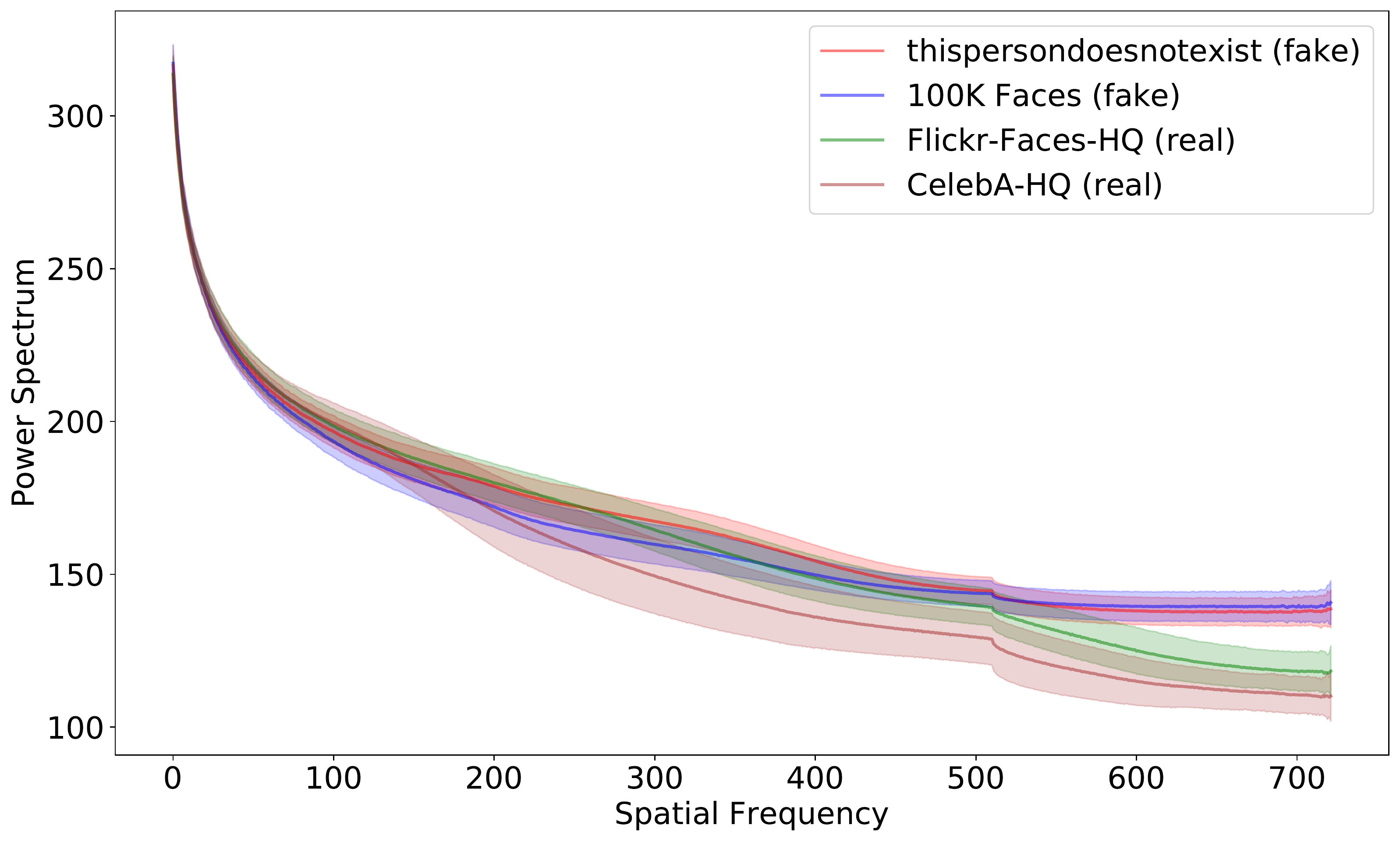}
   \caption{Statistics (mean and variance) of the Faces-HQ dataset.}
\label{fig:1000}
\end{figure}
\begin{figure}[!h]
\centering
   \includegraphics[width=\linewidth]{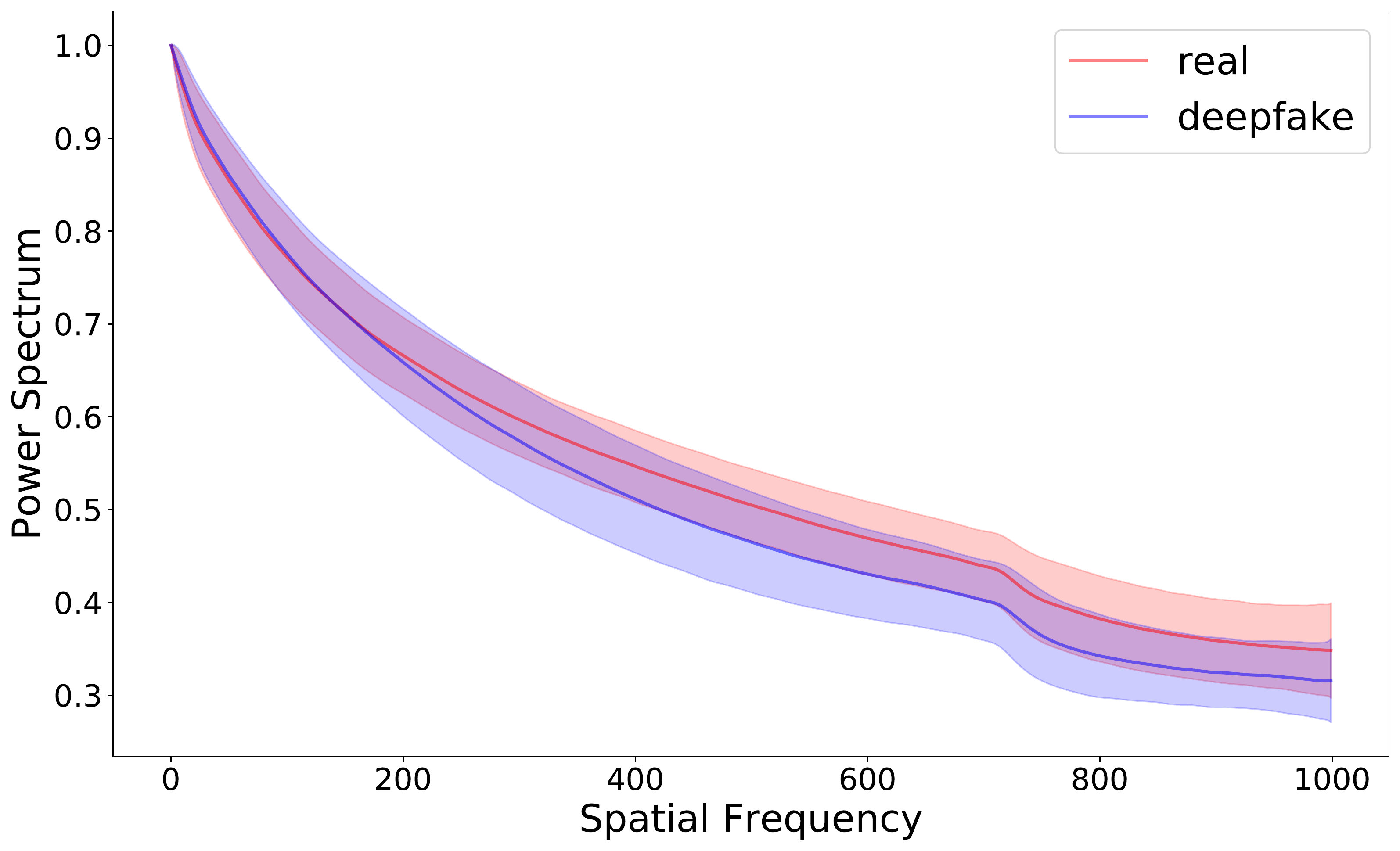}
   \caption{Statistics (mean and variance) of the FaceForensics++, DeepFakeDetection dataset.}
\label{fig:1000deepfak}
\end{figure}
\begin{figure}[!h]
\centering
   \includegraphics[width=\linewidth]{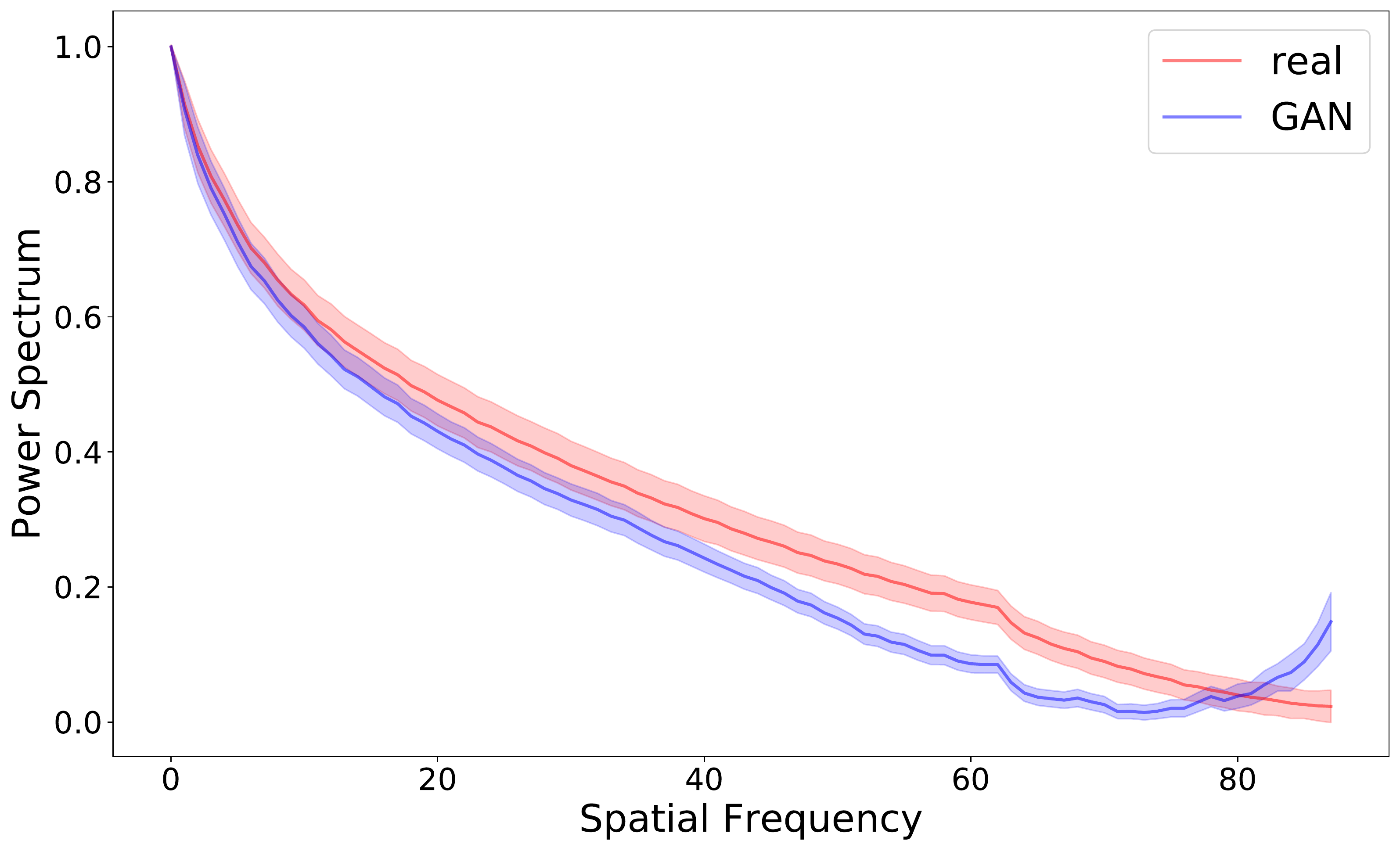}
   \caption{Statistics (mean and variance) of the CelebA dataset: average of images generated by the different GAN schemes (DCGAN, DRAGAN, LSGAN and WGAN-GP).}
\label{fig:1000gan}
\end{figure}

\begin{figure}[h]
\centering
  \includegraphics[width=\linewidth]{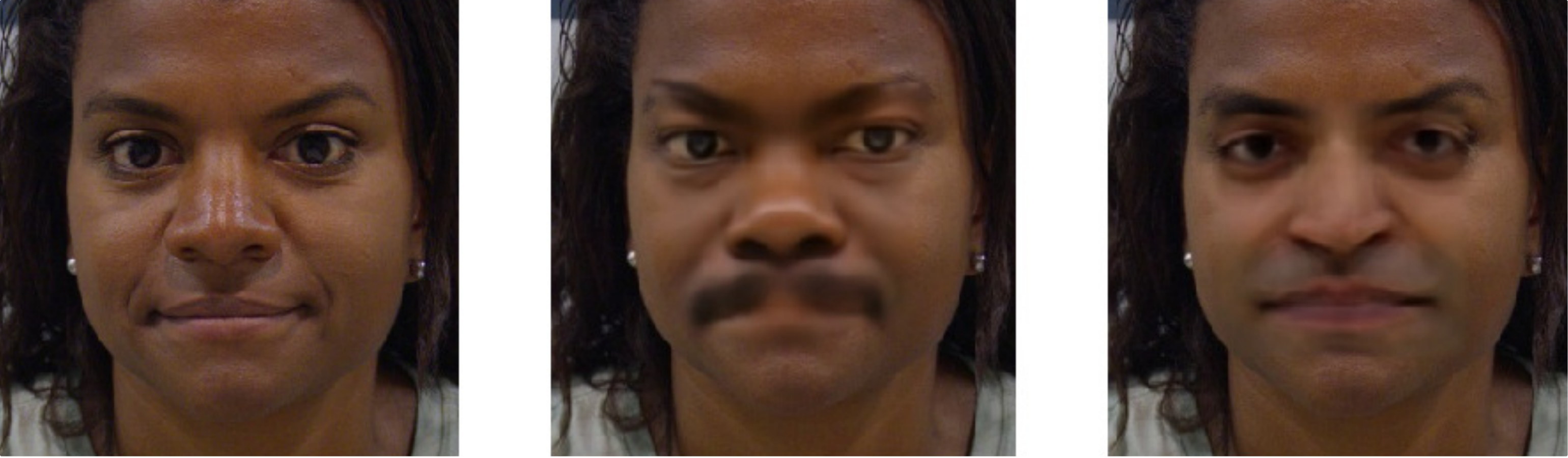}
  \\[2ex]
  \includegraphics[width=\linewidth]{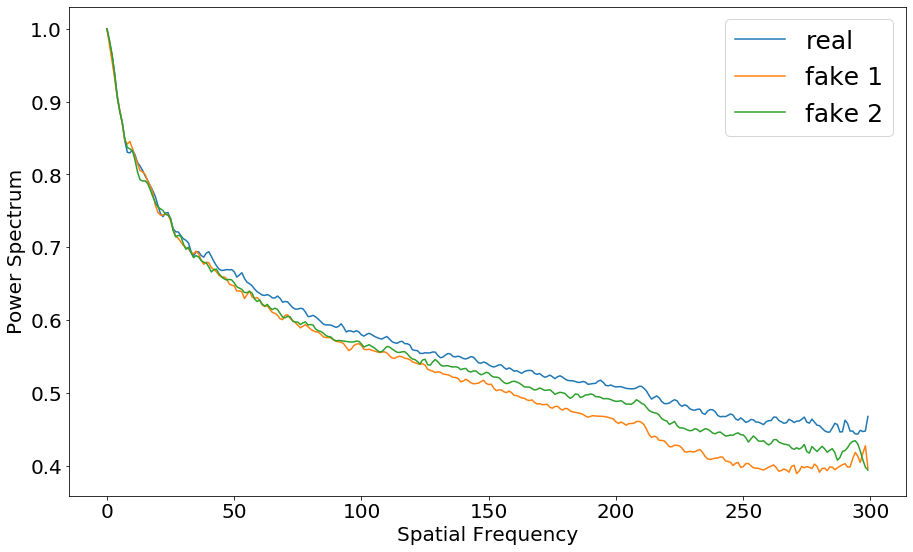}
  \caption{FaceForensics++ data. \textbf{Top:} example of one real face (left) and two deepfake faces, fake 1 (center) and fake 2 (right). Notice that the modifications only affect the inner face. \textbf{Bottom:} normalized and interpolated 1D Power Spectrum from the previous images.}
\label{fig:deepfaces}
\end{figure}

Figure \ref{fig:deepfaces} gives some additional data examples and their according spectral properties for the FaceForensics++ data.

\subsubsection{T-SNE Evaluation}
\begin{figure}[h]
\centering
   \includegraphics[width=\linewidth]{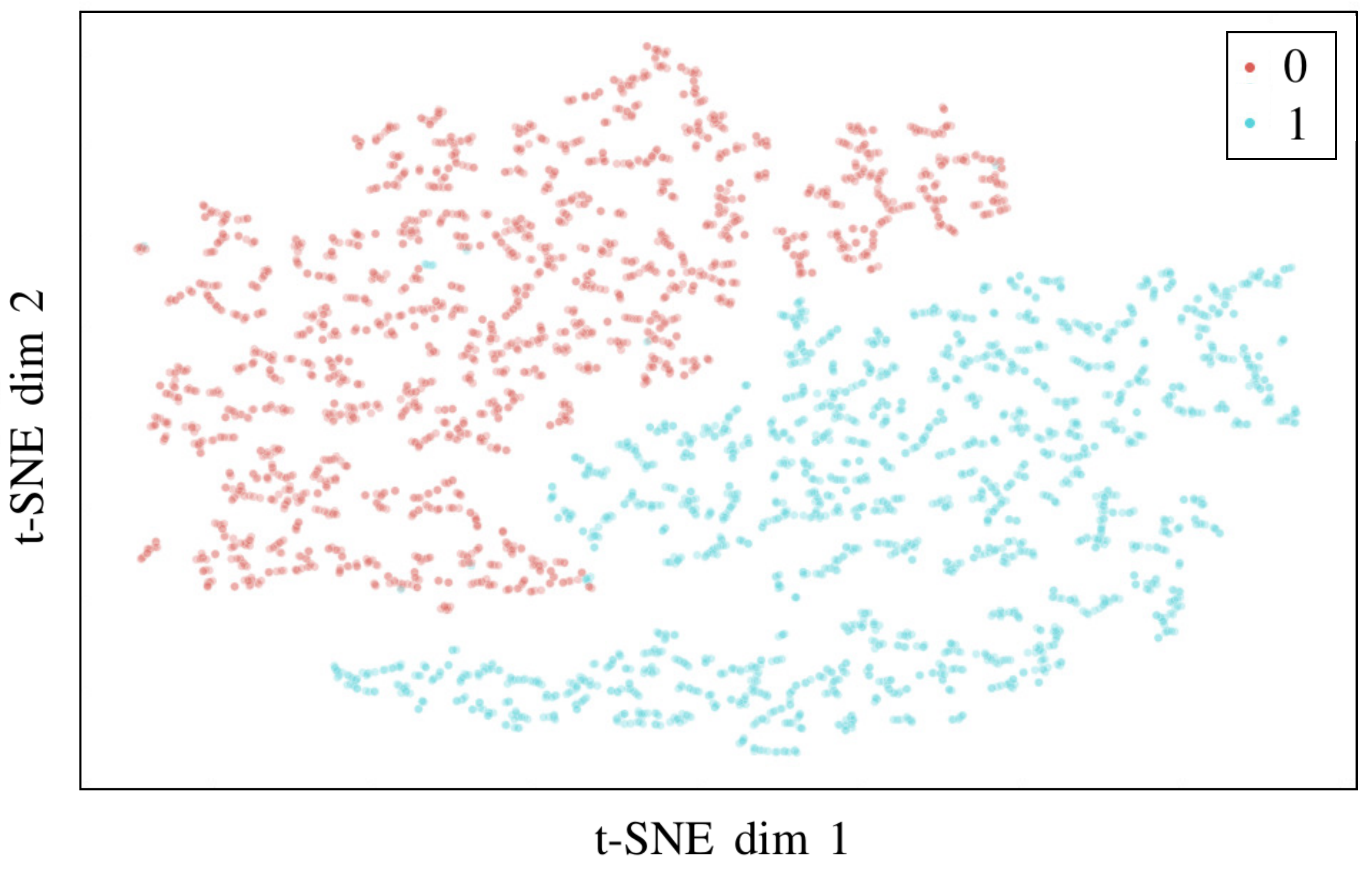}
	\caption{T-SNE visualization of 1D Power Spectrum on a random subset from \textit{Faces-HQ} data set. We used a perplexity of 4 and 4000 iterations to produce the plot.}
\label{fig:tsne}
\end{figure}

Figure \ref{fig:tsne} shows the clustering properties of our AI features. It is quite obvious that a classifier should not have problems to separate both classes (real and fake).

\subsubsection{Detection Results Depending on the Number of Available Samples}
In this section, we show some additional results on the DeepFake detection task (table 1 in the paper). In tables \ref{tab:both}, \ref{tab:both3} and \ref{tab:both2}, we focus on the effect of the available number of data samples during training. As shown in the paper, our approach works quite well in an unsupervised setting and needs as little as 16 annotated training samples to achieve 100\% classification accuracy in a supervised setting. 
\begin{table}[h]
\centering
\resizebox{0.35\textwidth}{!}{\begin{tabular}{c|ccc}
\hline
& \multicolumn{3}{c}{80\% (train) - 20\% (test)} \\
\cline{2-4}
\# samples & SVM & Logistic Reg. & K-Means  \\
\hline
4000 & 100\% & 100\% & 82\%  \\
1000 & 100\% & 100\% & 82\%  \\
100 & 100\% & 100\% & 81\%  \\
20 & 100\% & 100\% & 75\%  \\
\hline
\end{tabular}}
\caption{Faces-HQ: Test accuracy using SVM, logistic regression and k-means under different data settings.}
\label{tab:both}
\end{table}

\begin{table}[!h]
\centering
\resizebox{0.35\textwidth}{!}{\begin{tabular}{c|ccc}
\hline
& \multicolumn{3}{c}{80\% (train) - 20\% (test)} \\
\cline{2-4}
\# samples & SVM & Logistic Reg. & K-Means  \\
\hline
2000 & 100\% & 100\% & 96\%  \\
100 & 100\% & 95\% & 100\%  \\
20 & 100\% & 85\% & 100\%  \\
\hline
\end{tabular}}
\caption{CelebA: Test accuracy using SVM, logistic regression and k-means.}
\label{tab:both3}
\end{table}

\begin{table}[h]
\centering
\resizebox{0.3\textwidth}{!}{\begin{tabular}{c|cc}
\hline
& \multicolumn{2}{c}{80\% (train) - 20\% (test)} \\
\cline{2-3}
\# samples & SVM & Logistic Reg.\\
\hline
2000 & 85\% & 78\% \\
1000 & 82\% & 76\%  \\
200 & 77\% & 73\% \\
20 & 66\% & 76\% \\
\hline
\end{tabular}}
\caption{FaceForensics++: Test accuracy using SVM classifier and logistic regression classifier under different data settings. Evaluated on single frames.}
\label{tab:both2}
\end{table}

\section{Spectral Regularization on Auto-Encoder}
In this second section, we show some additional results from our AE experiments (see figure 4 of the paper).

\subsection{Loss during Training }
\begin{figure}[h]
\centering
   \includegraphics[width=\linewidth]{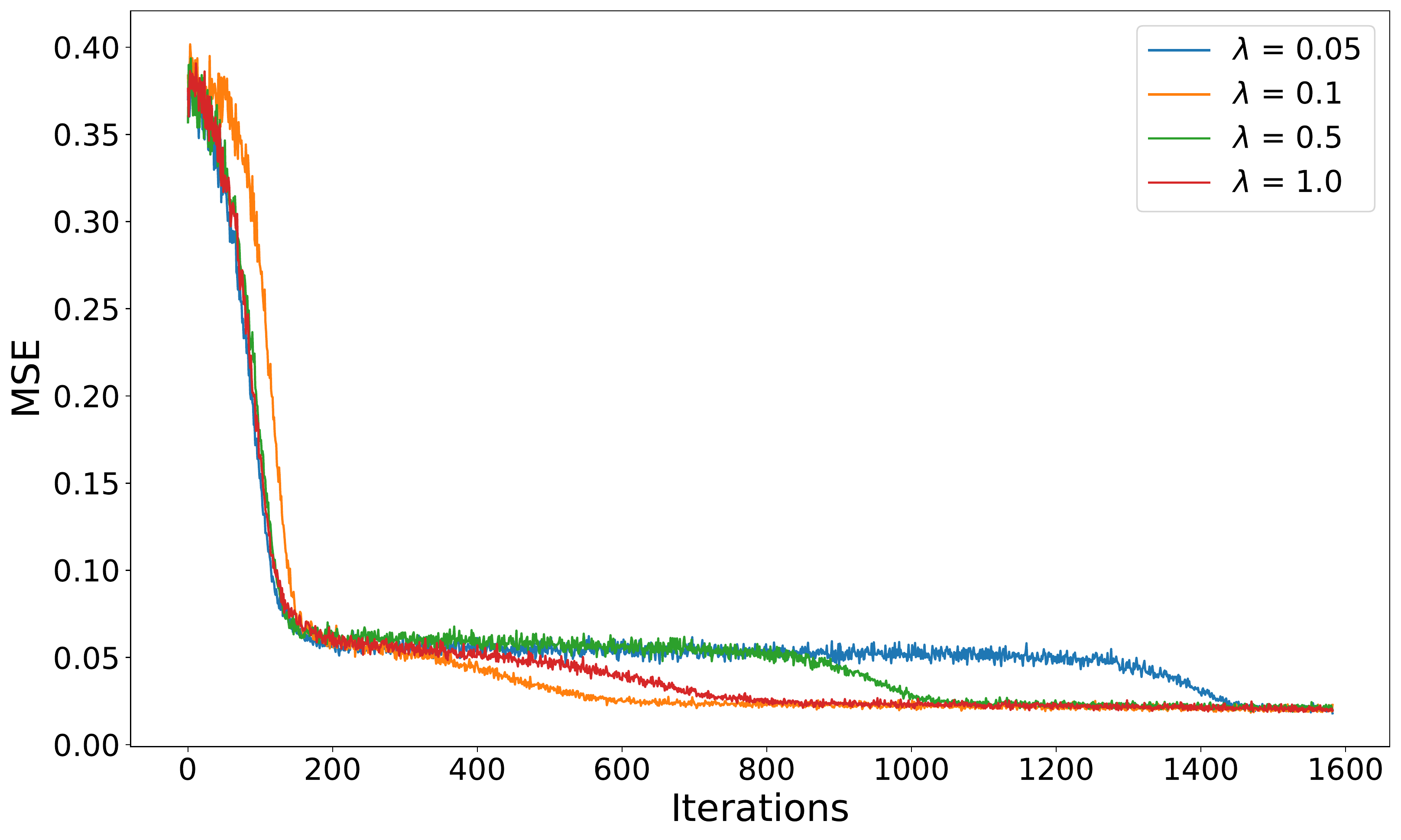}\\[2ex]
   \includegraphics[width=\linewidth]{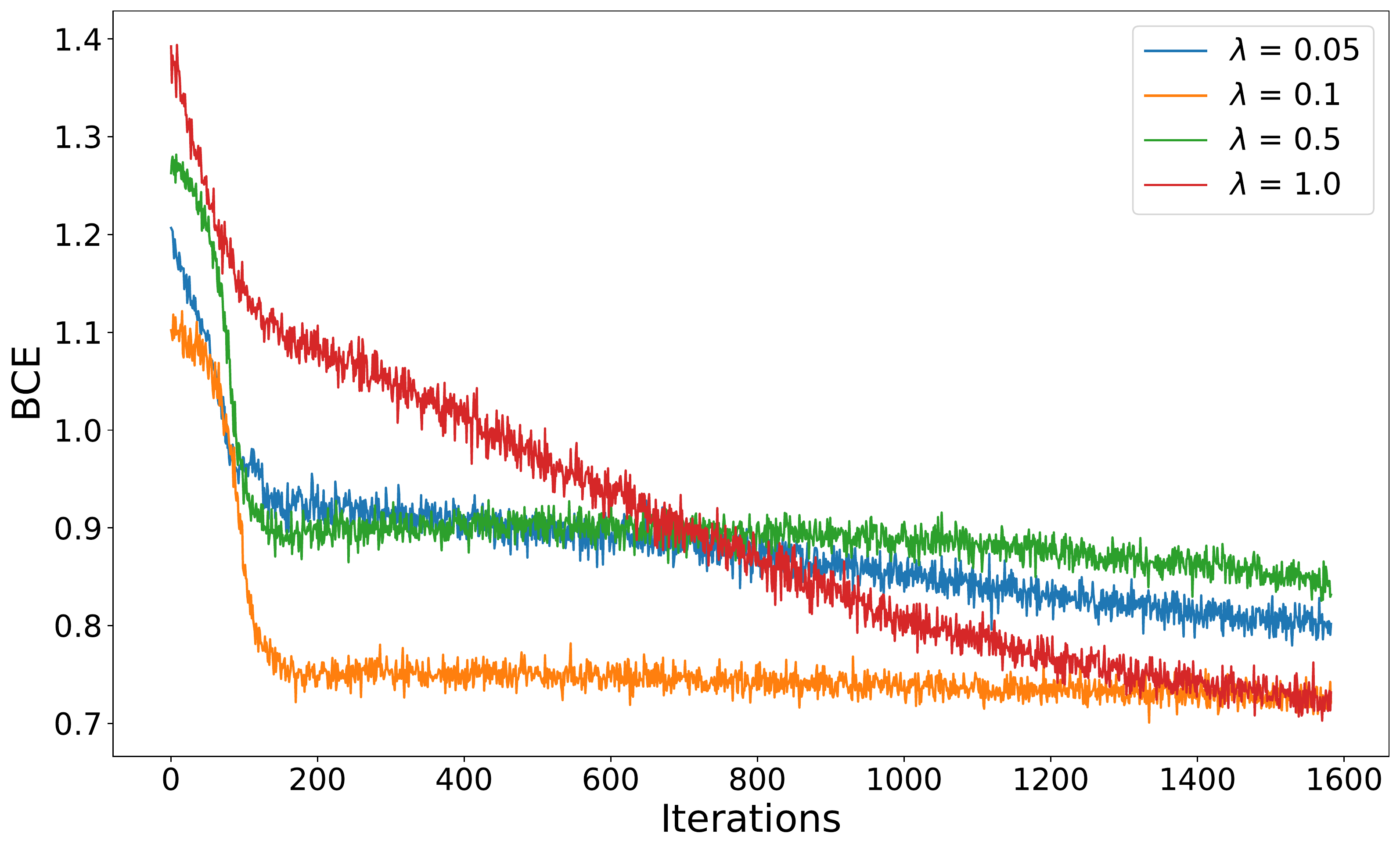}
	  \caption{Evolution of the different losses that define the $\mathcal{L}_{\mathrm{final}}$ from our AE. \textbf{Top:}  Mean Square Error (MSE) during the training ($\mathcal{L}_{\mathrm{Reconstruction}}$). \textbf{Bottom:}  Binary Cross-Entropy loss (BCE) during the training ($\mathcal{L}_{\mathrm{Spectral}}$).}
\label{fig:MSE}
\end{figure}

Figure \ref{fig:MSE} shows the evaluation of the loss (see equations 10 and 11 in the paper) with and without spectral regularization for a decoder with 3 convolutional layers and 3 filters of kernel size $5\times 5$ each.

These results show that the spectral regularization also has a positive effect on the convergence of the AE and the quality of the generated output images (in terms of MSE). 

\subsection{Effect of the Spectral Regularization}
\begin{figure}[t]
\centering
   \includegraphics[width=\linewidth]{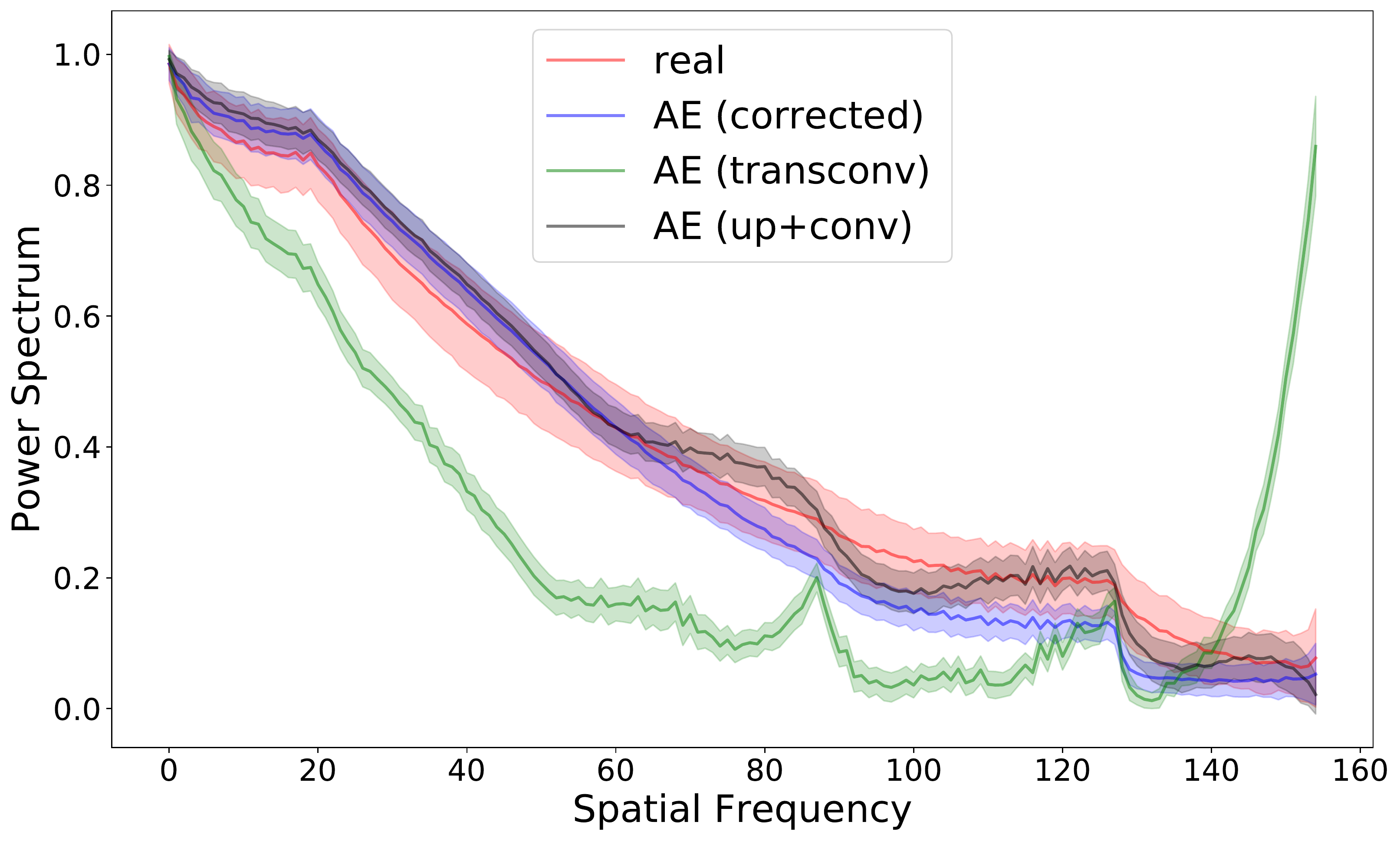}
	\caption{AE results for the baselines (\textit{transconv} and \textit{up+conv}) and for the proposal with spectral loss (\textit{corrected}). The corrected AE has 3 additional convolutional layers after the last \textit{transconv} layer. Each layer has 32 filters of size 5x5 and $\lambda=0.5$}
\label{fig:final}
\end{figure}

 Figure \ref{fig:final} shows the impact of the spectral regularization on the AE problem. We can notice how both \textit{transconv} and \textit{up+conv} suffer from different behaviour on the frequency spectrum domain, specially in high frequency components. Nevertheless, after applying our spectral regularization technique, the results get much closer to the real 1D Power Spectrum distribution, generating images closer to the real distribution.

\subsection{Effect of different Topologies}
\begin{figure}[!h]
\centering
   \includegraphics[width=\linewidth]{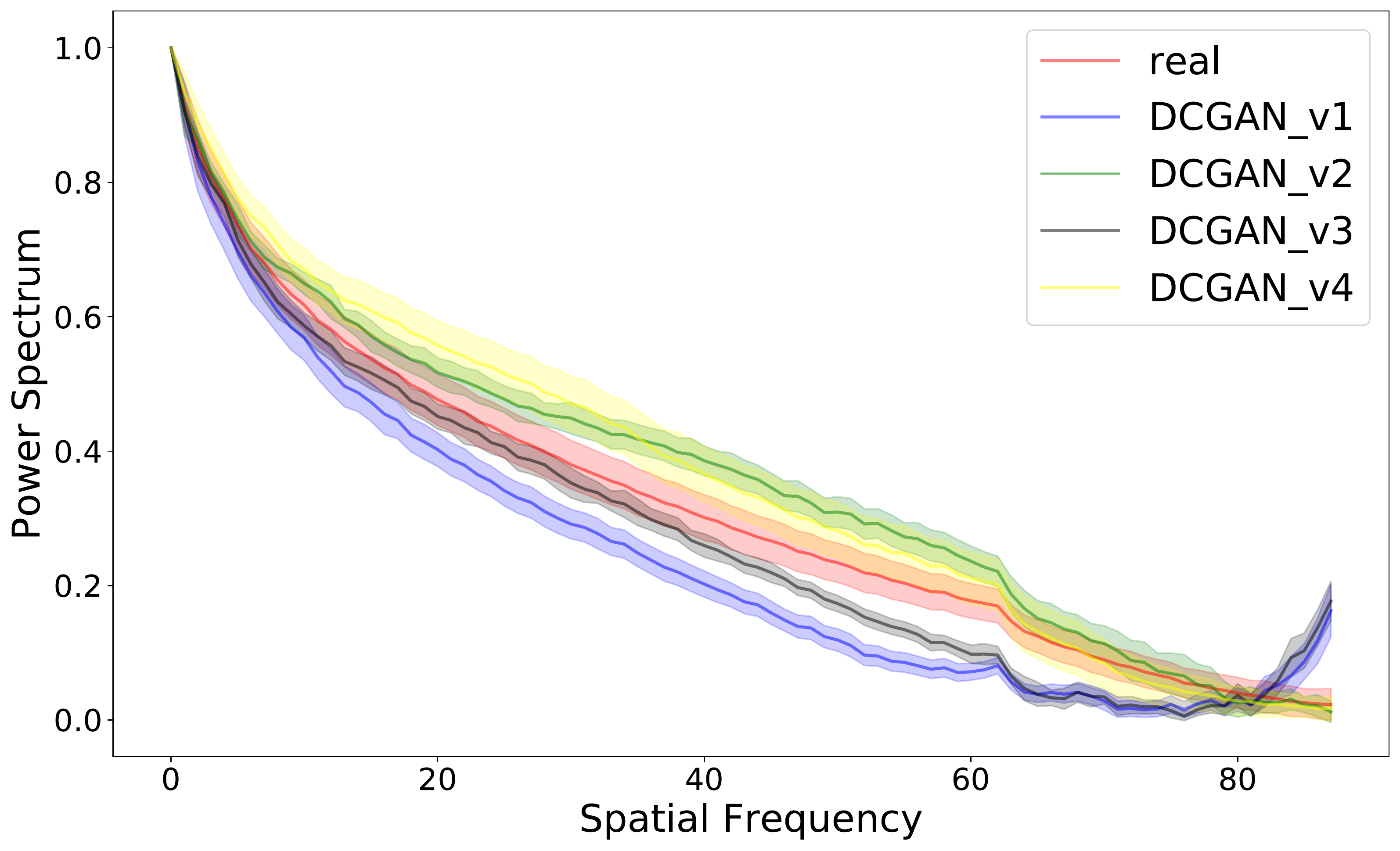}
   \caption{AE results for different topologies applied to DCGAN. Each version incorporates different amounts of convolutional layers to its DCGAN structure. }
\label{fig:diffGANdd}
\end{figure}

In this experiment, we evaluate the impact of different topology design choices. Figure \ref{fig:diffGANdd} shows statistics of the spectral distributions for some topologies:
\begin{itemize}
\item Real: original face images from CelebA
\item DCGAN\_v1: a DCGAN topology with spectral regularization and one convolution layer (32 5x5 filters) after the last two up-convolutions.
\item DCGAN\_v2: a DCGAN topology with spectral regularization and two convolution layers (32 5x5 filters) after the last up-convolution.
\item DCGAN\_v3: a DCGAN topology with spectral regularization and one convolution layer (32 5x5 filters) after the every up-convolution.
\item DCGAN\_v4: a DCGAN topology with spectral regularization and three convolution layers (32 5x5 filters) after the last up-convolution.
\end{itemize}
Following the theoretical analysis and after a rough topology search for verification, we conclude that it is sufficient to add 3 5x5 convolutional layers after the last up-convolution in order to utilize the spectral regularization. 

\end{document}